\documentclass[twoside,11pt]{article}

\usepackage[accepted]{melba}

\usepackage{amsmath,amsfonts}
\usepackage{xcolor}

\DeclareMathOperator*{\argmin}{arg\,min}

\usepackage[ruled, lined, linesnumbered, commentsnumbered, longend]{algorithm2e} 
\SetKwComment{Comment}{/* }{ */}

\usepackage{caption}
\usepackage{subcaption}
\usepackage{graphicx}

\definecolor{newcolor}{rgb}{.8,.349,.1}

\newcommand\redsout{\bgroup\markoverwith{\textcolor{red}{\rule[0.5ex]{2pt}{0.4pt}}}\ULon}

\definecolor{revisioncolor}{rgb}{0.,0.,0.0}
\newcommand{\revision}[1]{\textcolor{revisioncolor}{#1}}

\melbaheading{2022:007}{https://www.melba-journal.org/papers/2022:007.html}{2022}{1--27}{09/2021}{03/2022}{Perkonigg, Hofmanninger, Herold, Prosch, and Langs}{Information Processing in Medical Imaging (IPMI) 2021}{Aasa Feragen, Stefan Sommer, Mads Nielsen, Julia Schnabel}
\ShortHeadings{Continual Active Learning Using Pseudo-Domains}{Perkonigg, Hofmanninger, Herold, Prosch, and Langs}
\firstpageno{1}

\title{Continual Active Learning Using Pseudo-Domains for Limited Labelling Resources and Changing Acquisition Characteristics}

\author{\name Matthias Perkonigg \email matthias.perkonigg@meduniwien.ac.at \\
\addr Department of Biomedical Imaging and Image-guided Therapy, Computational Imaging Research Lab (CIR), Medical University of Vienna, Austria
	\AND
	\name Johannes Hofmanninger \\
	\addr Department of Biomedical Imaging and Image-guided Therapy, Computational Imaging Research Lab (CIR), Medical University of Vienna, Austria
    \AND
    \name Christian Herold \\
    \addr Department of Biomedical Imaging and Image-guided Therapy, Medical University of Vienna, Austria
    \AND
    \name Helmut Prosch \\
    \addr Department of Biomedical Imaging and Image-guided Therapy, Medical University of Vienna, Austria
	\AND
	\name Georg Langs \email georg.langs@meduniwien.ac.at \\
	\addr Department of Biomedical Imaging and Image-guided Therapy, Computational Imaging Research Lab (CIR), Medical University of Vienna, Austria
}

\begin{document}

\maketitle

\begin{abstract}
Machine learning in medical imaging during clinical routine is impaired by changes in scanner protocols, hardware, or policies resulting in a heterogeneous set of acquisition settings. When training a deep learning model on an initial static training set, model performance and reliability suffer from changes of acquisition characteristics as data and targets may become inconsistent. Continual learning can help to adapt models to the changing environment by training on a continuous data stream. However, continual manual expert labelling of medical imaging requires substantial effort. Thus, ways to use labelling resources efficiently on a well chosen sub-set of new examples is necessary to render this strategy feasible. 
Here, we propose a method for continual active learning operating on a stream of medical images in a multi-scanner setting. The approach automatically recognizes shifts in image acquisition characteristics -- new \emph{domains} --, selects optimal examples for labelling and adapts training accordingly. Labelling is subject to a limited budget, resembling typical real world scenarios. \revision{In order to avoid catastrophic forgetting while learning on new domains the proposed method utilizes a rehearsal memory.} To demonstrate generalizability, we evaluate the effectiveness of our method on three tasks: cardiac segmentation, lung nodule detection and brain age estimation. \revision{Results show that the proposed approach outperforms other active learning methods on a continuous data stream with domain shifts.}
\end{abstract}


\section{Introduction}
\begin{figure}[t]
    \centering
    \includegraphics[width=0.9\textwidth]{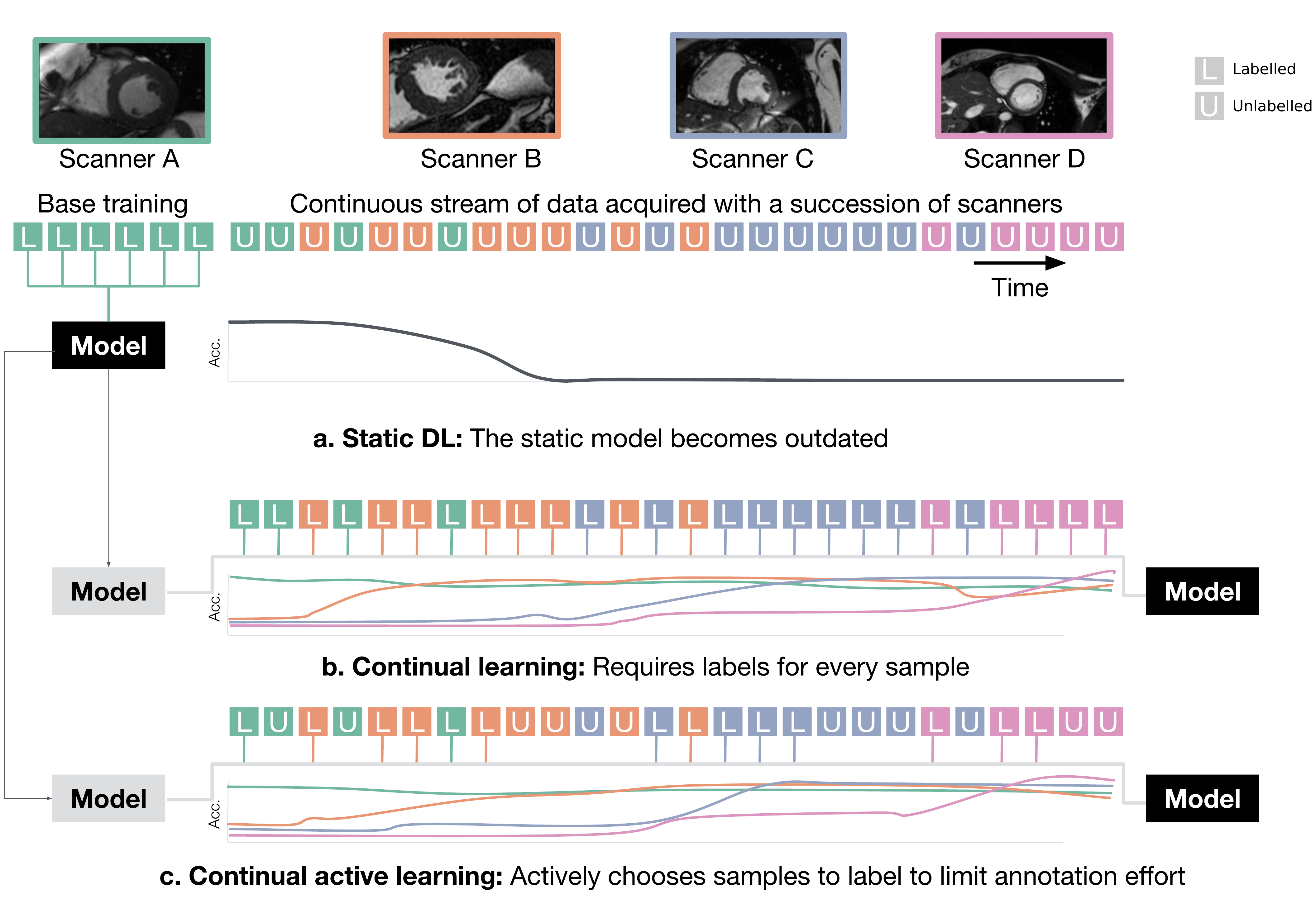}
    \caption{Experimental setup: A model is pre-trained on scanner A data (base training) and then subsequently updated on a data stream gradually including data of scanner B, C and D. (a) The accuracy of a model trained on a static data set of only scanner A drops as data from other scanner appear in the data stream. (b) Continual learning can incorporate new knowledge, but requires all samples in the data stream to be labelled. (c) Active continual learning actively chooses the labels to annotated from the stream and is able to keep the model up to date while limiting the annotation effort.}
    \label{fig:exp_setup}
\end{figure}

The performance of deep learning models in the clinical environment is hampered by frequent changes in scanner hardware, imaging protocols, and heterogeneous composition of acquisition routines. Ideally, models trained on a large data set should be continuously adapted to the changing characteristics of the data stream acquired in imaging departments. However, training on a data stream of images acquired solely by recent acquisition technology can lead to \textit{catastrophic forgetting} \citep{McCloskey1989}, a deterioration of performance on preceding domains or tasks, see Figure \ref{fig:exp_setup} (a). Therefore, a continual learning strategy is required to counteract forgetting. \revision{Counteracting forgetting is important in medical imaging to ensure backward compatibility of the model, as well as to enable faster adaptation to possibly related domains in the future.} Model training in a medical context requires expert labelling of data in new domains. This is often prohibitively expensive and time-consuming. Therefore, reducing the number of cases requiring labelling, while still providing training with the variability necessary to generalize well, is a key challenge in active learning on medical images~\citep{Budd2019}. Here, we propose an active learning approach to make efficient use of annotation resources during continual machine learning. In a continual data stream of examples from an unlabelled distribution, it identifies those that are most informative if labelled next. 

We focus on accounting for domain shifts occurring in a continual data stream, without knowledge about when those shifts occur. Figure \ref{fig:exp_setup} depicts the scenario our method is designed for. A deep learning model is trained on a base data set of labelled data of one domain (Scanner A), afterwards it is exposed to the data stream in which scanners B, C and D occur. For each sample of the data stream, continual active learning has to take a decision on whether or not labelling is required for the given image. Labelled images are then used for continual learning with a rehearsal memory. Previously proposed continual active learning methods either disregard domain shifts in the training distribution or assume that the domain membership of images is known \citep{Lenga2020, Ozgun2020, Karani2018}. However, this knowledge can not be assumed in clinical practice due to the variability in encoding the meta data \citep{Gonzalez2020}. Therefore, a technique to detect those domain shifts in a continuous data stream is needed.
A combination of continual active learning with automatic detection of domain shifts is desirable to ensure that models can deal with a diverse and growing number of image acquisition settings, and at the same time minimizing the manual efforts and resources needed to keep the models up to date. 

\paragraph{Contribution}  
Here, we propose an continual active learning method. The approach operates without domain membership knowledge and learns by selecting informative samples to annotate from a continuous data stream. We first run a base training on data of a single scanner. Subsequently, the continuous data stream is observed and domain shifts in the stream are detected. This detection triggers the labelling of samples of the newly detected pseudo-domain and knowledge about the new samples is incorporated to the model. At the same time, the model should not forget knowledge about previous domains, thus we evaluate the final model on data from all observed domains. Our approach combines continual active learning with a novel domain detection method for continual learning. We refer to our approach as \textit{Continual Active Learning for Scanner Adaptation (CASA)}. CASA uses a rehearsal method to alleviate catastrophic forgetting and an active labelling approach without prior domain knowledge. CASA is designed to learn on a continuous stream of medical images under the restriction of a labelling budget, to keep the required manual annotation effort low. The present paper expands on our prior work on continual active learning~\citep{Perkonigg2021ContinualAcquisition}. \revision{In this prior work we introduced the novel setup on active and continual learning on a data stream of medical imaging and proposed the CASA method. Here, we expand the approach in several ways: (1) The pseudo-domain assignment is refined and simplified. While previous work used a method based on isolation forests \citep{Liu2008}, in this work we use a distance metric for pseudo-domain assignment. (2) Experiments with two additional machine learning tasks, cardiac segmentation in MR imaging and lung nodule detection in CT are included to demonstrate the generalizability of CASA. (3) Active learning with uncertainty is added as a reference method across all experiments to compare the performance of CASA. (4) A more detailed analysis of the composition of the rehearsal memory, and the influence of the sequential nature of the data stream are included.}

\section{Related Work}
The performance of machine learning models can be severely hampered by changes in image acquisition settings \citep{Castro2020, Glocker2019MachineEffects, Prayer2021VariabilityStudy}. \textit{Harmonization} can counter this in medical imaging~\citep{Fortin2018HarmonizationSites, Beer2020LongitudinalData}, but requires all data to be available at once, a condition not feasible in an environment where data arrives continually. 
\emph{Domain adaptation (DA)} addresses domain shifts, and in particular approaches  dealing with continuously shifting domains are related to the proposed method. \cite{Wu2019} showed how to adapt a machine learning model for semantic segmentation of street scenes under different lightning conditions. Rehearsal methods for domain adaptation have been shown to perform well on benchmark data sets such as rotated MNIST \citep{Bobu2018AdaptingDomains} or Office-31 \citep{Lao2020}. In the area of medical imaging, DA is used to adapt between different image acquisition settings \cite{Guan2021DomainSurvey}. However, similar to harmonization, most DA methods require that source and target domains are accessible at the same time. 
\revision{}

\emph{Continual learning (CL)} is used to incorporate new knowledge into ML models without forgetting knowledge about previously seen data. For a detailed review on CL see \citep{Parisi2019ContinualReview, Delange2021ATasks}. An overview of the potential of CL combined with medical imaging combined is given in \citep{Pianykh2020}. \cite{Ozdemir2018} used continual learning to incrementally add new anatomical regions into a segmentation model. Related to this work, CL has been used for domain adaptation for chest X-ray classification \citep{Lenga2020} and for brain MRI segmentation \citep{Ozgun2020}. \cite{Karani2018} proposed a simple, yet effective approach for lifelong learning for brain MRI segmentation by using separate batch normalization layers for each protocol. 
The rehearsal memory approach of our work is closely related to \emph{dynamic memory}, a continual learning method based on an image style-based rehearsal memory \citep{Hofmanninger2020a, Perkonigg2021DynamicImaging}. However dynamic memory assumes a fully labelled data set, while this approach limits the annotation need by using an active stream-based selective sampling method.

\emph{Active Learning} is an area of research where the goal is to identify samples to label next to minimize annotation effort, while maximizing training efficiency. A detailed review of active learning in medical imaging is given in \citep{Budd2019}. In context of this review our work is closest related to \textit{stream-based selective sampling}.  Also \cite{Pianykh2020} discuss human-in-the-loop concepts with continual learning, which is similar to the approach presented in this work.  Active learning was used to classify fundus and histopathological images by \cite{Smailagic2020} in an incremental learning setting. \cite{Zhou2021ActiveEfforts} combine transfer learning and active learning to choose samples for labelling based on entropy and diversity. They show the benefits of their method on polyp detection and pulmonary embolism detection. Different from the proposed method, those approaches do not take data distribution shifts during training into account and do not perform selective sampling based on a continuous data stream.

\section{Methods}
The continual active learning method CASA uses a \textit{rehearsal memory} and performs active training sample selection from an unlabelled, continuous data stream $\mathcal{S}$ to keep a task model up-to-date under the presence of domain shifts, while at the same time countering forgetting. For active sample labelling an oracle can be queried to return task annotations $y=\mathbf{o}(x)\ |\ x\in\mathcal{S}$. In a real world clinical setting this oracle can be thought of as a radiologist. Due to the cost of manual labelling, the queries to the oracle are limited by the labelling budget $\beta$. CASA aims at training a task network on a continuous data stream under the restriction of $\beta$, while at the same time alleviating catastrophic forgetting. CASA detects \textit{pseudo-domains} to keep a diverse set of training samples in the rehearsal memory. Those pseudo-domains are formed by groups of examples with similar appearance. Similarity is measured as style difference of images. The proposed method consists of a pseudo-domain module, a task module and and two memories (outlier memory and training rehearsal memory), that are controlled by the CASA-Algorithm described in the following.

\begin{figure}[t]
    \centering
    \includegraphics[width=0.8\textwidth]{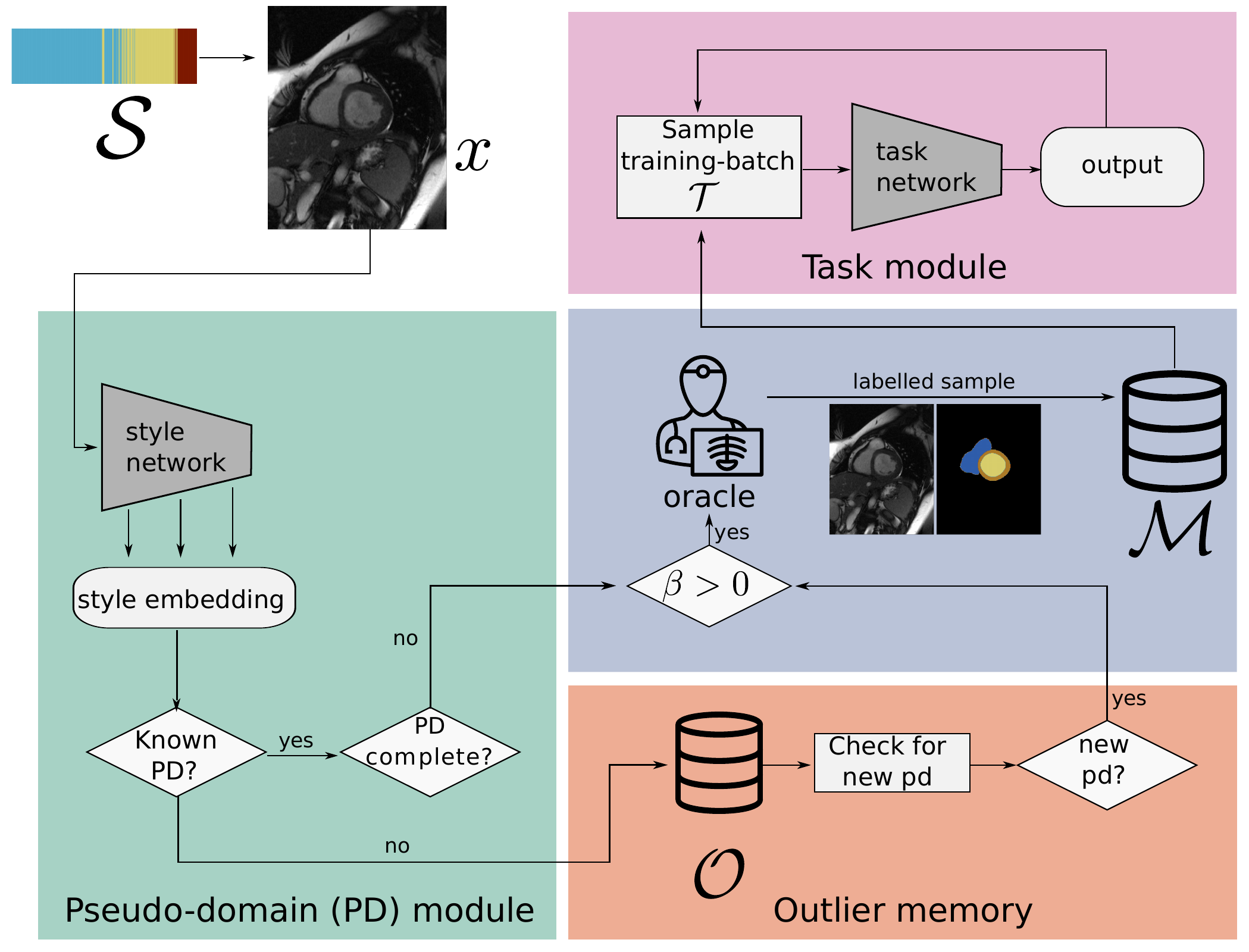}
    \caption{Overview of the \textit{CASA} algorithm. Each sample from the data stream is processed by the pseudo-domain module to decide whether its routed to the oracle or to the outlier memory. Whenever a new item is added to the outlier memory  it is evaluated if a new pseudo-domain (pd) should be created. The oracle labels a sample and stores it in the training memory, from which the task module trains a network. Binary decision alternatives resulting in discarding the sample are left out for clarity of the figure.}
    \label{fig:CASA_overview}
\end{figure}

\subsection{CASA Training Scheme}

Before starting continual training, the task module is pre-trained on a labelled data set $\mathcal{L} = \{\langle i_1,l_1\rangle, \dots, \langle i_L,l_L\rangle\}$ of image-label pairs $\langle i, l \rangle$ obtained on a particular scanner (\textit{base training}). \revision{This base training is a conventional epoch based training procedure assuming a static, fully labelled data set. In this training phase samples can be revisited in a supervised training scheme without restriction of the labelling budget.}
After base training is finished, continual training is started from a model which performs well on data of a single scanner. Continual active training follows the scheme depicted in Figure \ref{fig:CASA_overview} and outlined in Algorithm~\ref{algo:casa}.
First, an \textit{input-mini-batch} $\mathcal{B} = \{x_1, \dots, x_B\}$ is drawn from $\mathcal{S}$ and the \textit{pseudo-domain module} evaluates the style embedding (see Section \ref{sec:style_emb}) of each image $x\in \mathcal{B}$. Based on this embedding, a decision is taken to store $x$ in one of the memories ($\mathcal{O}$ or $\mathcal{M}$) or discard $x$. The fixed sized \textit{training memory} $\mathcal{M} = \{\langle m_1,n_1,d_1\rangle \dots, \langle m_M,n_M, d_M\rangle\}$, where $m$ is the image, $n$ the corresponding label and $d$ the assigned \textit{pseudo-domain}, holds samples the task network can be trained on. Labels $n$ can only be generated by querying the \textit{oracle} $\mathbf{o}(x)$, and are subject to the limited labelling budget $\beta$. $\mathcal{M}$ is initialized with a random subset of $\mathcal{L}$ before starting continual training. 
Pseudo-domain detection is performed within the \textit{outlier memory} $\mathcal{O} = \{\langle o_1,c_1\rangle \dots, \langle o_n,c_n\rangle\}$, which holds a set of unlabelled images. $o$ represents the image and $c$ is a counter, how long the image is part of $\mathcal{O}$. Details about the outlier memory are given in Section \ref{sec:outlier_memory}.
Given that training has not saturated on all pseudo-domains, a training step is performed by sampling \textit{training-mini-batches} $\mathcal{T} = \{\langle t_1,{u}_1\rangle, \dots, \langle t_T,u_T\rangle\}$ of size $T$ from $\mathcal{M}$, and training the task module (Section \ref{sec:task_module}) for one step. This process is continued by drawing the next mini-batch $\mathcal{B}$ from $\mathcal{S}$.

\begin{algorithm}
	\caption{CASA Training Algorithm}\label{algo:casa}
	\SetKwFunction{styleembedding}{styleembedding}
	\SetKwFunction{pdassignment}{pd-assignment}
	\SetKwFunction{add}{add}
	\SetKwFunction{pdcomplete}{pd-complete}
	\SetKwFunction{sample}{sample}
	\SetKwFunction{train}{train}
	\SetKwFunction{nextfu}{nextBatch}
	\SetKwFunction{newdomaincheck}{newPseudodomainCheck}
	\SetKwFunction{trainingfinished}{trainingFinished}
    \SetKwInOut{KwIn}{Input}
    \SetKwInOut{KwOut}{Output}
    \KwIn{Pre-trained task model $t$, continual data stream $\mathcal{S}$, limited budget $\beta$, task-memory $\mathcal{M}$, outlier memory $\mathcal{O}$, $b$ training steps per batch}
		\While{$\mathcal{B}\gets\nextfu(\mathcal{S})$}{
		    \For{$x\in\mathcal{B}$}{
		       $e\gets \styleembedding(x)$ \\
		       $pd\gets \pdassignment(e)$ \\
		       \If{$pd==-1$}{
		          $\mathcal{O}.\add(x)$}
		       \Else{
		            \If{\pdcomplete(pd)}{
    		            \If{$\beta>0$}{
    		                $\mathcal{M}.\add(x)$\\
    		                $\beta\gets \beta-1$
    		            }
		            }
		        }
		   }
		   $\mathcal{N} \gets \newdomaincheck(\mathcal{O}, \beta)$ \Comment*[r]{elements of discovered pd}
		   \If{$\mathcal{N} \neq \emptyset$}{
		      \For{$n\in\mathcal{N}$}{
		         \If{$\beta>0$}{
		                $\mathcal{M}.\add(x)$\\
		                $\beta\gets \beta-1$
    		     }
		      }
		   }
		        \For{$i\gets0$ \KwTo $b$}{
    		        $\mathcal{T}\gets\sample(\mathcal{M})$\\
    		        $\train(t, \mathcal{T})$
		        }
		}
\end{algorithm}

\subsection{Pseudo-domain module}
\label{sec:style_emb}
CASA does not assume direct knowledge about domains (e.g. scanner vendor, scanning protocol). Therefore, in the pseudo-domain module the \textit{style} of each image $x$ is evaluated and $x$ is assigned to a pseudo-domain. Pseudo-domains represent groups of images which exhibit similar style. 
A set of pseudo-domains $\mathcal{D} = \{\langle c_1, d_1, \bar{p}_1 \rangle \dots \langle c_D, d_D, \bar{p}_D\rangle\}$ is defined by their \textit{style embedding} center $c_j$ and the maximum distance $d_j$ from $c_j$ that an image is considered belonging to pseudo-domain $j$. In addition, a running average $\bar{p}_j$ of the performance on $j$ for each pseudo-domain $j\in \{1, \dots, D\}$ is stored.

\paragraph{Style embedding} A style embedding is calculated for an image $x$ based on a \textit{style network} pre-trained on a different dataset (not necessarily related to the task) and not updated during training. \revision{The choice of the style network is dependent on the dataset used for training, the specific style networks used in this paper are discussed in Section \ref{sec:exp_setup}}. From this network, we evaluate the style of an image based on the gram matrix $G^l \in \mathbb{R}^{N_l \times N_l}$, where $N_l$ is the number of feature maps in layer $l$. Following \citep{Gatys2016,Hofmanninger2020a}, $G_{ij}^l(x)$ is defined as the inner product between the vectorized activations $\mathbf{f}_{il}(x)$ and $\mathbf{f}_{jl}(x)$ of two feature maps $i$ and $j$ in a layer $l$ given a sample image $x$:
\begin{equation}
\displaystyle G_{ij}^l(x) = \frac{1}{N_lM_l} \mathbf{f}_{il}(x)^\top\mathbf{f}_{jl}(x)
\label{eq:grammatrix}
\end{equation}
where $M_l$ denotes the number of elements in the vectorized feature map. 
Based on the gram matrix a \textit{style embedding} $\mathbf{e}(x)$ is defined: For a set of convolutional layers $\mathcal{C}$ of the style network, gram matrices ($G^l\ |\ l \in \mathcal{C}$) are calculated and Principle Component Analysis (PCA) is applied to reduce the dimensionality of the embedding to a fixed size of $e$. PCA is fitted on style embeddings of the base training set.  

\paragraph{Pseudo-domain assignment} CASA uses pseudo-domains to assess if training for a specific style is needed and to diversify the memory $\mathcal{M}$. A new image $x\in \mathcal{B}$ is assigned to the pseudo-domain minimizing the distance between the center of the pseudo-domain and the style embedding $\mathbf{e}(x)$ according to the following equation:
\begin{equation}
    \mathbf{p}(x) = 
\begin{cases}
    \underset{d}{\argmin}\ |\mathbf{e}(x)-c_d| & \text{if}\ |\mathbf{e}(x)-c_d|<d_d\\
    -1,              & \text{otherwise}
\end{cases}
|\ d \in \mathcal{D}
\end{equation}

If $\mathbf{p}(x)=-1$, the image is added to the outlier memory $\mathcal{O}$ from which new pseudo-domains are detected (see Section \ref{sec:outlier_memory}).
\revision{For the threshold distance $d_d$, let $\mathcal{M}_d$ be the subset of samples assigned to domain $d$. Then $d_d$ is calculated as two times the mean distance between the center of $d$ and the style embedding of all samples in $\mathcal{M}_d$:
\begin{equation}
    d_d =  2 \cdot \frac{\sum_{x \in \mathcal{M}_d} (c_d-\mathbf{e}(x))^2}{|\mathcal{M}_d|}
\end{equation}
}
If the pseudo-domain $\mathbf{p}(x)$ is known and has completed training, we discard the image, otherwise it is added to $\mathcal{M}$ according to the strategy described in Section \ref{sec:training_memory}.

\paragraph{Average performance metric} $\bar{p}_j$ is the running average of a performance metric of the target task calculated on the last $P$ elements of pseudo-domain $j$ that have been labelled by the \textit{oracle}. The performance metric is measured before training on the sample. $\bar{p}_j$ is used to evaluate if the pseudo-domain completed training, that is $\bar{p}_{j}>k$ for classification tasks and $\bar{p}_{j}<k$ for regression tasks, where $k$ is a fixed performance threshold. \revision{If that threshold is reached, subsequent samples assigned to the corresponding pseudo-domain do not require manual annotation.} \revision{The specific choice of the performance metric depends on the learning task, see Section \ref{sec:exp_setup} for the metrics used in the experiments of this work.}

\subsection{Task module}
\label{sec:task_module}
The task module is responsible for learning the target task (e.g. cardiac segmentation), where the main component of this module is the \textit{task network} ($\mathbf{t}(x)\mapsto y$), mapping from input image $x$ to target label $y$. During base training, this module is trained on a labelled data set $\mathcal{L}$. During continual active training, the module is updated in every step by drawing $n$ training-input-batches $\mathcal{T}$ from the memory $\mathcal{M}$ and performing a training step on each of the batches. The aim of CASA is to train a task module performing well on images of all image acquisition settings available in $\mathcal{S}$ without suffering catastrophic forgetting.

\subsection{Training memory}
\label{sec:training_memory}
The $M$ sized training memory $\mathcal{M}$ is balanced between the pseudo-domains currently in $\mathcal{D}$. Each of the $D$ pseudo-domains can occupy up to $\frac{M}{D}$ elements in the memory. If a new pseudo-domain is added to $\mathcal{D}$ (see Section \ref{sec:outlier_memory}) a random subset of elements of all previous domains is flagged for deletion, so that only $\frac{M}{D}$ elements are kept protected in $\mathcal{M}$. 
If a new element $e=\langle m_k, n_k, d_k \rangle$ is inserted to $\mathcal{M}$ and $\frac{M}{D}$ is not reached, an element currently flagged for deletion is replaced by $e$. Otherwise the element will replace the one in $\mathcal{M}$, which is of the same pseudo-domain and minimizes the distance between the style embeddings. Formally, the element with index $\xi$ is replaced:
\begin{equation}
 \xi(i) = \argmin_{j}(\mathbf{e}(m_k)- \mathbf{e}(m_j))^2 |\ n_k = n_j,\ j \in \{1,\dots,M\}.
\end{equation}

\subsection{Outlier memory and pseudo-domain detection}
\label{sec:outlier_memory}
The outlier memory $\mathcal{O}$ holds candidate examples that do not fit an already identified pseudo-domain, and might form a new pseudo-domain by themselves. Whether they form a pseudo-domain is determined based on their proximity in the style embedding space. Examples are stored until they are assigned a new pseudo-domain, or if a fixed number of training steps $z$ is reached. If no new pseudo-domain is discovered for an image within $z$ steps, it is considered a 'real' outlier and removed from the outlier memory. Within $\mathcal{O}$, new pseudo-domains are discovered, and subsequently added to $\mathcal{D}$. The discovery process is started when $\lvert \mathcal{O}\rvert=o$, where $o$ is a fixed threshold of minimum elements in $\mathcal{O}$. To detect a dense region in the style embedding space of samples in the outlier memory, the pairwise euclidean distances of all elements in $\mathcal{O}$ are calculated. If there is a group of images for which all pair-wise distances are below a threshold $t$, a new pseudo-domain is established by these images. For all elements belonging to the new pseudo-domain, labels are queried from the oracle and they are transferred from $\mathcal{O}$ to $\mathcal{M}$.

\section{Experimental Setup}

We evaluate CASA on data streams containing imaging data sampled from different scanners. To demonstrate the generalizability of CASA to a range of different areas in medical imaging, three different tasks are evaluated: 
\begin{itemize}
    \item \textit{Cardiac segmentation} on cardiovascular magnetic resonance (CMR) data 
    \item \textit{Lung nodule detection} in computed tomography (CT) images of the lung 
    \item \textit{Brain Age Estimation} on T1-weighted MRI data
\end{itemize}

For all tasks, the performance of CASA is compared to several baseline techniques (see Section \ref{sec:baselines}). 

\subsection{Data set}

\begin{table}
\begin{subtable}[c]{\textwidth}
\centering
    \begin{tabular}{l|l|l|l|l|l}
        \centering
 &  Siemens (C1) &     GE (C2) &  Philips (C3) &  Canon (C4) & Total \\ \hline\hline

Base  &   1120 &    0 &      0 &    0  & 1120\\ 
Continual &    614 &  720 &   2206 &  758  & 4298\\ 
Validation   &    234 &  248 &    220 &  258  & 960\\ 
Test  &    228 &  246 &    216 &  252 & 942 \\\hline
        \hline
    \end{tabular}
    \subcaption{Cardiac segmentation data set}
    \label{tbl:data_cardiac}
\end{subtable}
\begin{subtable}[c]{\textwidth}
    \centering
    \begin{tabular}{l|l|l|l|l|l}
 &    GE/L (L1) &    GE/H (L2) &   Siemens (L3) &  LNDb (L4) & Total\\\hline\hline
Base  &  253 &    0 &    0 &           0 & 253 \\
Continual &  136 &  166 &  102 &         479  & 883\\
Validation   &   53 &   23 &   10 &          55  & 141 \\
Test  &   85 &   26 &   18 &          91 & 220\\\hline
        \hline
    \end{tabular}
    \subcaption{Lung nodule detection data set}
    \label{tbl:data_lung}
\end{subtable}
\begin{subtable}[c]{\textwidth}
    \centering
    \begin{tabular}{l|l|l|l|l|l}
        \centering

         &   1.5T IXI (B1) & 1.5T OASIS (B2) & 3.0T IXI (B3) &  3.0T OASIS (B4) &  Total \\
\hline \hline
Base            &   201 &      0 &      0 & 0 &  201 \\
Continual       &   52 &   190 & 146 &   1504 &   1892\\
Validation       &    31 & 23 &    18 &    187 &   259\\
Test             &    31 & 23 &    18 &    187 &   259\\
        \hline
    \end{tabular}
    \subcaption{Brain age estimation data set}
\end{subtable}
\caption{Splitting of the data sets into a base training, continual training, validation, and test set. The number of cases in each split are shown.}
\label{tbl:data}
\end{table}

\paragraph{Cardiac segmentation} 2D cardiac segmentation experiments were performed on data from a multi-center, multi-vendor challenge data set \citep{Campello2021Multi-CentreChallenge}. The data set included CMR data from four different scanner vendors (Siemens, General Electric, Philips and Canon), where we considered each vendor as a different domain. We split the data into base training, continual training, validation, and test set on a patient level. Table \ref{tbl:data} (a) shows the number of slices for each domain in those splits. Manual annotations for left ventricle, right ventricle and left ventricular myocardium were provided. 2D images were center-cropped to 240$\times$196px and normalized to a range of [0-1]. In the continual data set, the scanners appeared in the order Siemens, GE, Philips and Canon and are referred to Scanner C1-C4. 

\paragraph{Lung nodule detection} For lung nodule detection, we used two data sources: the LIDC-database \citep{Armato2011}, with the annotations as provided for the LUNA16-challenge \citep{Setio2017} and the LNDb challenge data set \citep{Pedrosa2019}. 
Lung nodule detection was performed as 2D bounding box detection, therefore bounding boxes around annotated lesions were constructed for all available lung nodule annotations.
From LIDC, the three most common domains, in terms of scanner vendor and reconstruction kernel, were used to collect a data set suitable for continual learning with shifting domains. Those domains were GE MEDICAL SYSTEMS with low frequency reconstruction algorithm (GE/L), GE MEDICAL SYSTEM with high frequency reconstruction algorithm (GE/H) and Siemens with B30f kernel (Siemens). In addition, data from LNDb was used as a forth domain which was comprised of data from multiple Siemens scanners. For LNDb, nodules with a diameter $<3mm$ were excluded to match the definition in LIDC. Image intensities were cropped from -1024 to 1024 and normalized to [0-1].
From the images 2D slices were extracted and split into base training, continual training, validation and test data set according to Table \ref{tbl:data} (b). For all continual learning experiments, the order of the domains was GE/L, GE/H, Siemens and LNDb, those are referred to as L1-L4. 

\paragraph{Brain age estimation}
Data pooled from two different data sets containing three different scanners was used for brain age estimation. The IXI data set\footnote{https://brain-development.org/ixi-dataset/} and data from OASIS-3 \citep{LaMontagne2019} was used to collect a continual learning data set. From IXI, we used data from a Philips Gyroscan Intera 1.5T and a Philips Intera 3.0T scanner, from OASIS-3 we used data from a Siemens Vision 1.5T and a Siemens TrioTim 3.0T scanner. Images were resized to 64x128x128px and normalized to a range between 0 and 1. Data was split into base base training, continual training, validation and test set (see Table \ref{tbl:data} (c)). In continual training data occurred in the order: Philips 1.5T, Siemens 1.5T, Philips 3.0T and Siemens 3.0T, the scanner domains are referred to B1-B4 in the following.

\subsection{Experimental setup}
\label{sec:exp_setup}
\revision{
\paragraph{Hyperparameters}
Multiple hyperparamters are used in CASA. In the experiments the focus is to analyze those parameters with the most influence on the methods performance are the memory size $M$ and the labelling budget $\beta$. These two parameters are extensively evaluated and analyzed in Section \ref{sec:res_memsize} and \ref{sec:res_budget}. 
Besides that the dimensions of the style embedding after PCA is fixed to $e=30$ and the minimal number of elements in $\mathcal{O}$ to discover new pseudo-domains is fixed to $o=10$ after preliminary experiments showed little influence on model performance of those parameters. Another hyperparameter is the choice of the style network that is fixed during training, here preliminary experiments showed that the choice of the style network is not critical for the performance of CASA.
The performance threshold $k$ depends on the task and performance metric used and is set to be approximately as high as the average performance of domain specific models (see Section \ref{sec:baselines}). The threshold used $t$ is dependent on the data set used, therefore the threshold was empirically set by analyzing mean distances of the style embeddings of the base training data set. Details on $k$ and $t$ are given in the following.}

\paragraph{Cardiac segmentation} For segmentation, a 2D-UNet \citep{Ronneberger2015} was used as task network. The style network was a ResNet-50 \citep{Ren2017}, pretrained on ImageNet and provided in the torchvision package. For segmentation, the performance metric used in all experiments was the mean dice score (DSC) over the three annotated anatomical regions (left ventricle, right ventricle and left ventricular myocardium). \revision{The performance threshold $k$ is fixed to a mean DSC of 0.75 based on domain specific models. The distance threshold was fixed to $t=0.025$.}

\paragraph{Lung nodule detection} As a task network, Faster R-CNN with a ResNet-50 backbone was used \citep{Ren2017}. For evaluating the style, we used a ResNet-50 pretrained on ImageNet. For lung nodule detection, we used average precision (AP) as the performance metric to evaluate the performance of the models with a single metric. We followed the AP definition by \cite{Everingham2010}. \revision{For lung nodule detection we set $k$ to an AP of 0.5 and $t=0.025$ for all experiments.}

\paragraph{Brain age estimation} As a task network, a simple 3D feed-forward network was used \citep{Dinsdale2020}. The style network used in the pseudo-domain module was a 3D-ModelGenesis model, pre-trained on computed tomography images of the lung \citep{Zhou2020}. \revision{For brain age estimation a 3D data set was used, thus we used a different style model as for cardiac segmentation and lung nodule detection.} The main performance measure for brain age estimation we used was the mean absolute error (MAE) between predicted and true age. \revision{For brain age estimation we set $k$ to a MAE to 5.0 and $t=0.025$ for all experiments.}

\subsection{Methods compared}
\label{sec:baselines}
Throughout the experiments, five methods were evaluated and compared:
\begin{enumerate}
    \item \textit{Joint model (JM)}: a model trained in a standard, epoch-based approach on samples from all scanners in the experiment jointly.
    \item \textit{Domain specific models (DSM)}: a separate model is trained for each domain in the experiment with standard epoch-based training. The evaluation for a domain is done for each domain setting separately.
    \item \textit{Naive AL (NAL)}: a naive continuously trained, active learning approach of labelling every $n$-th label from the data stream, where $n$ depends on the labelling budget $\beta$.
    \item \textit{Uncertainty AL (UAL)}: Is a common type of active learning which labels samples where the task network is uncertain about the output \citep{Budd2019}. Here, uncertainty is calculated using dropout at inference as an approximation for Bayesian inference \citep{Gal2016}.
    \item \textit{CASA (proposed method)}: The method described in this work. 
\end{enumerate}

Joint models and DSM require the whole training data set to be labelled, and thus are an upper limit to which the continual learning methods are compared to. CASA, UAL and NAL use an oracle to label specific samples only. The comparison to NAL and UAL evaluates if the detection of pseudo-domains and labelling based on them is beneficial in an active learning setting.  Note, that the aim of our experiments is to show the gains of CASA compared to other active learning methods, not to develop new state-of-the-art methods for either of the three tasks evaluated.

\subsection{Experimental evaluation}
We evaluate different aspects of CASA:
\begin{enumerate}
    \item \textbf{Performance across domains:} For all tasks, we evaluate the performance across domains at the end of training, and highlight specific properties of CASA in comparison to the baseline methods. Furthermore, we evaluate the ability of continual learning to improve accuracy on existing domains by adding new domains \textit{backward transfer} (BWT), and the contribution of previous domains in the training data to improving the accuracy on new domains \textit{forward transfer} (FWT) \citep{Lopez-Paz2017}. BWT measure how learning a new domain influences the performance on previous tasks, FWT quantifies the influence on future tasks. Negative BWT values indicate catastrophic forgetting, thus avoiding negative BWT is especially important for continual learning.
    \item \textbf{Influence of labelling budget $\beta$:} For cardiac segmentation, the influence of the $\beta$ is studied. The labelling budget is an important parameter in clinical practice, since labelling new samples is expensive. We express $\beta$ as a fraction of the continual data set. Different settings of $\beta$ are analysed $\beta=\frac{1}{5}$, $\beta=\frac{1}{8}$, $\beta=\frac{1}{10}$ and $\beta=\frac{1}{20}$. To solely study the influence of $\beta$, the memory size in this experiments is fixed $M=128$ for all settings.
    \item \textbf{Influence of memory size $M$:} For cardiac segmentation different settings for the memory size $M$ are evaluated. The memory size is the number of samples that are stored for rehearsal, and might be limited due to privacy concerns and/or storage space. Here, $M$ is evaluated for $[64, 128, 256, 512, 1024]$ and a fixed $\beta=\frac{1}{10}$. \revision{We assume that the diversity of the data addressed by CASA is a result of the set of scanners, not the number of images in the dataset, therefore $M$ is fixed to specific numbers rather than a fraction of the dataset.}
    \item \textbf{Memory composition and pseudo-domains:} We study if our proposed method of detecting pseudo-domains is keeping samples in memory that are representative of the whole training set for cardiac segmentation. In addition, we evaluate how the detected pseudo-domains are connected to the real domains determined by the scanner types.
    \item \textbf{Learning on a random stream:} We study how CASA is performing on a random stream of data, where images of different acquisition settings are appearing randomly in the data stream. In contrast to the standard setting, where these acquisition settings appear subsequently with a phase of transition in between.
\end{enumerate}

\section{Results}
\label{sec:results}
\subsection{Performance across domains}
\label{sec:quantiative_results}
Here, the quantitative results at the end of the continual training are compared for a memory size $M=128$ and a labelling budget of $\beta=\frac{1}{10}$. Different settings for $M$ and $\beta$ are evaluated in Section \ref{sec:res_budget} and \ref{sec:res_memsize} respectively.

\paragraph{Cardiac segmentation} Performance for cardiac segmentation was measured using the mean dice score. Continual learning with CASA applied to cardiac segmentation outperformed UAL and NAL for scanners C2, C3 and C4 (Table \ref{tbl:res_cardiac}). For scanner C1, the performance of the model trained with CASA was slightly below UAL and NAL. This was due to the distribution in the rehearsal memory, where CASA balanced between all four scanner domains, while for UAL and NAL, a majority of the rehearsal memory was filled with C1 images (further details are discussed in Section \ref{sec:memory_analysis}). Compared to the base model, which corresponds to the model performance prior to continual training the performance of CASA remained constant for C1 and at the same time rose significantly for C2 ($+0.041$), C3 ($+0.085$) and C4 ($+0.336$), showing that CASA was able to perform continual learning without forgetting the knowledge acquired in base training. UAL and NAL were also able to learn without forgetting during continual learning, this is also reflected in a BWT of around $0$ for all compared methods. However, UAL and NAL performed worse in terms of FWT and overall dice for C2 to C4.
As expected, JModel outperformed all other training strategies since it has access to the fully labelled training set at once and thus can perform epoch-based deep learning. 
\begin{table}[h]
\centering
\small
\resizebox{\textwidth}{!}{
\begin{tabular}{|l|l|l|l|l||l|l|}
\hline
Meth.   & C1 & C2 & C3 & C4 & BWT & FWT\\
\hline
CASA  & $0.812\pm0.017$ & $0.731\pm0.025$ & $0.803\pm0.015$ & $0.676\pm0.158$ & $-0.006\pm0.009$ & $0.086\pm0.046$\\
UAL & $0.816\pm0.006$ & $0.700\pm0.009$ & $0.764\pm0.023$ & $0.652\pm0.078$ & $-0.003\pm0.013$ & $0.067\pm0.031$\\
NAL &  $0.819\pm0.003$ & $0.707\pm0.005$ & $0.761\pm0.013$ & $0.564\pm0.064$ & $-0.004\pm0.003$ & $0.060\pm0.026$\\
\hline \hline
DSM & $0.835\pm0.047$ & $0.718\pm0.018$ & $0.773\pm0.016$ & $0.833\pm0.003$\\
JModel & $0.828\pm0.009$ & $0.758\pm0.020$ & $0.818\pm0.023$ & $0.825\pm0.016$ \\ 
Base & $0.814$& $0.690$ & $0.718$ & $0.340$\\ 
\cline{1-5}
\end{tabular}
}
\caption{Cardiac segmentation: Quantitative results for $M=128$, $\beta=\frac{1}{10}$ measured in mean dice score. $\pm$ marks the standard deviations over $n=5$ independent training runs. \revision{Comparison between CASA (proposed method), Uncertainty AL (UAL), Naive AL (NAL), Domain specific models (DSM), Joint Model (JModel) and the model after base training. C1-C4 denote the scanners occuring in the continuous data stream. BWT and FWT mark backward and forward transfer respectively.} }
\label{tbl:res_cardiac}
\end{table}

\paragraph{Lung nodule detection}
In Table \ref{tbl:res_lung}, results for lung nodule detection measured as average precision are compared. CASA performed significantly better than NAL and UAL for all scanners. For L4, which were the images extracted from LNDb, the distribution of nodules was different. For scanners L1-L3, the mean lesion diameter was 8.29mm, while for L4, lesion diameter was 5.99mm on average. This lead to a worse performance on L4 for all approaches. Nevertheless, CASA was the only active learning approach able to label a large enough amount of images for L4 such that it can significantly outperform the base model, as well as NAL and UAL.

\begin{table}[h]
\centering
\small
\resizebox{\textwidth}{!}{
\begin{tabular}{|l|l|l|l|l||l|l|}
\hline
Meth.   & L1 & L2 & L3 & L4 & BWT & FWT\\
\hline
CASA  & $0.664\pm0.016$ & $0.543\pm0.080$ & $0.816\pm0.005$ & $0.229\pm0.026$ & $0.023\pm0.037$  & $0.025\pm0.036$\\
UAL & $0.650\pm0.011$ & $0.394\pm0.058$ & $0.738\pm0.038$ & $0.180\pm0.021$ & $-0.003\pm0.048$ & $-0.015\pm0.023$ \\
NAL &  $0.619\pm0.025$ & $0.472\pm0.057$ & $0.765\pm0.041$ & $0.184\pm0.019$ & $-0.019\pm0.025$ & $0.004\pm0.009$ \\
\hline \hline
DSM & $0.644\pm0.036$ & $0.440\pm0.060$ & $0.488\pm0.102$  & $0.365\pm0.062$ \\
JModel & $0.728\pm0.033$ & $0.649\pm0.037$  & $0.793\pm0.017$  & $0.454\pm0.024$  \\ 
Base & $0.644$& $0.458$ & $0.807$ & $0.159$\\ 
\cline{1-5}
\end{tabular}
}
\caption{Lung nodule detection: Quantitative results for $M=128$, $\beta=\frac{1}{10}$ measured in average precision. $\pm$ marks the standard deviations over $n=5$ independent training runs. \revision{Comparison between CASA (proposed method), Uncertainty AL (UAL), Naive AL (NAL), Domain specific models (DSM), Joint Model (JModel) and the model after base training. L1-L4 denote the scanners occuring in the continuous data stream. BWT and FWT mark backward and forward transfer respectively.}}
\label{tbl:res_lung}
\end{table}

\paragraph{Brain age estimation} Table \ref{tbl:res_brain} (c) shows the results for brain age estimation in terms of MAE. CASA was able to perform continual learning without forgetting, and outperformed UAL and NAL for all scanners (B1-B4) at the end of the continuous data stream. Comparing MAE for B1 data for UAL ($7.01$) and NAL ($11.91$) with the base model ($6.44$) shows that forgetting has occurred for UAL and NAL. For CASA, MAE for B2 and B3 was notably higher than for B1 and B4 respectively. Due to the composition of the continual training set, B2 (n=190) and B3 (n=146) occurred less than B4 (n=1504) in the data stream, consequently leading to fewer B2 and B3 images seen during training, and consequently a worse performance. Nevertheless, CASA was able to handle this data set composition better than UAL and NAL. 
\begin{table}[h]
\centering
\small
\begin{tabular}{|l|l|l|l|l||l|l|}
\hline
Meth.   & B1 & B2 & B3 & B4 & BWT & FWT\\
\hline
CASA  & $6.40\pm0.35$ & $8.96\pm1.16$ & $8.56\pm1.14$ & $6.54\pm0.70$ &  $0.45\pm0.59$ & $4.97\pm0.23$\\
UAL & $7.01\pm0.68$  & $12.22\pm1.78$ & $9.75\pm0.50$ & $12.92\pm1.47$ & $1.16\pm0.39$ & $1.66\pm0.52$ \\
NAL &  $11.91\pm2.31$ & $17.67\pm2.90$ & $14.16\pm3.52$ & $15.54\pm2.20$ & $1.87\pm2.36$ & $2.82\pm3.44$\\
\hline \hline
DSM & $6.28\pm0.37$ & $4.77\pm0.57$ & $7.16\pm0.53$  & $4.42\pm1.13$ \\
JModel & $6.51\pm1.07$ & $6.63\pm2.09$ & $4.38\pm1.31$ & $5.99\pm0.53$  \\ 
Base & $6.44$& $18.26$ & $11.43$ & $15.86$\\ 
\cline{1-5}
\end{tabular}
\caption{Brain Age Estimation: Quantitative results for $M=128$, $\beta=\frac{1}{10}$ measured in mean absolute error. $\pm$ marks the standard deviations over $n=5$ independent training runs. \revision{Comparison between CASA (proposed method), Uncertainty AL (UAL), Naive AL (NAL), Domain specific models (DSM), Joint Model (JModel) and the model after base training. B1-B4 denote the scanners occuring in the continuous data stream. BWT and FWT mark backward and forward transfer respectively.}}
\label{tbl:res_brain}
\end{table}

\subsection{Influence of labelling budget $\beta$}
\label{sec:res_budget}
\begin{figure}[t]
    \centering
    \includegraphics[width=0.9\textwidth]{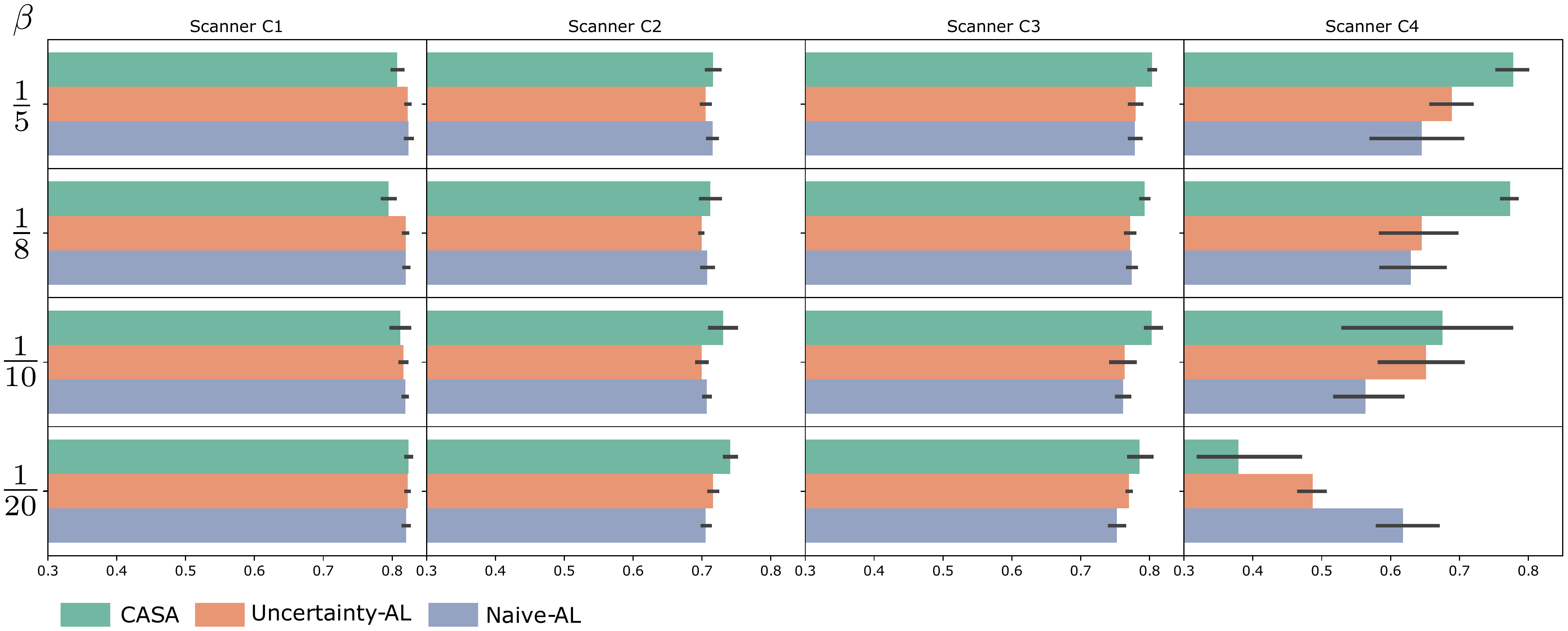}
    \caption{Influence of labelling budget $\beta$ for cardiac segmentation with $M=128$ comparing CASA, uncertainty AL and naive AL. Performance was measured in mean DSC.}
    \label{fig:label_budget}
\end{figure}

The influence of $\beta$ on cardiac segmentation performance is shown in Figure \ref{fig:label_budget}. For the first scanners C1 and C2, a similar performance can be observed for all methods and values of $\beta$. This was due to the fact that all methods have a sufficient amount of budget to adapt from C1 to C2. For C3, the performance for CASA was slightly higher compared to UAL and NAL. The most striking difference between the methods can be seen for scanner C4. There, CASA performed significantly better for $\beta=\frac{1}{5}$ and $\beta=\frac{1}{8}$. For $\beta=\frac{1}{10}$ CASA outperformed UAL and NAL on average however, a large deviation between the individual five runs is observable. Investigating this further revealed that CASA ran out of budget before C4 data appeared in the stream for one of the random seeds. Thus it was not able to adapt to C4 for this random seed. For $\beta=\frac{1}{20}$, CASA consumed the whole labelling budget before C4 data occured in the stream. Thus, it was not able to adapt to C4 data properly. UAL only had little budget left when C4 data came in and runs out afterwards thus, the performance was significantly worse to the results with more labelling budget. NAL performed the best for the setting with the lowest budget $\beta=\frac{1}{20}$, due to the fact that NAL labels every 20th step and thus did not run out of budget until the end of the stream is reached. 

\subsection{Influence of memory size $M$}
\label{sec:res_memsize}
For cardiac segmentation, we investigate the influence of the training memory size $M$. 
\begin{figure}[t]
    \centering
    \includegraphics[width=0.9\textwidth]{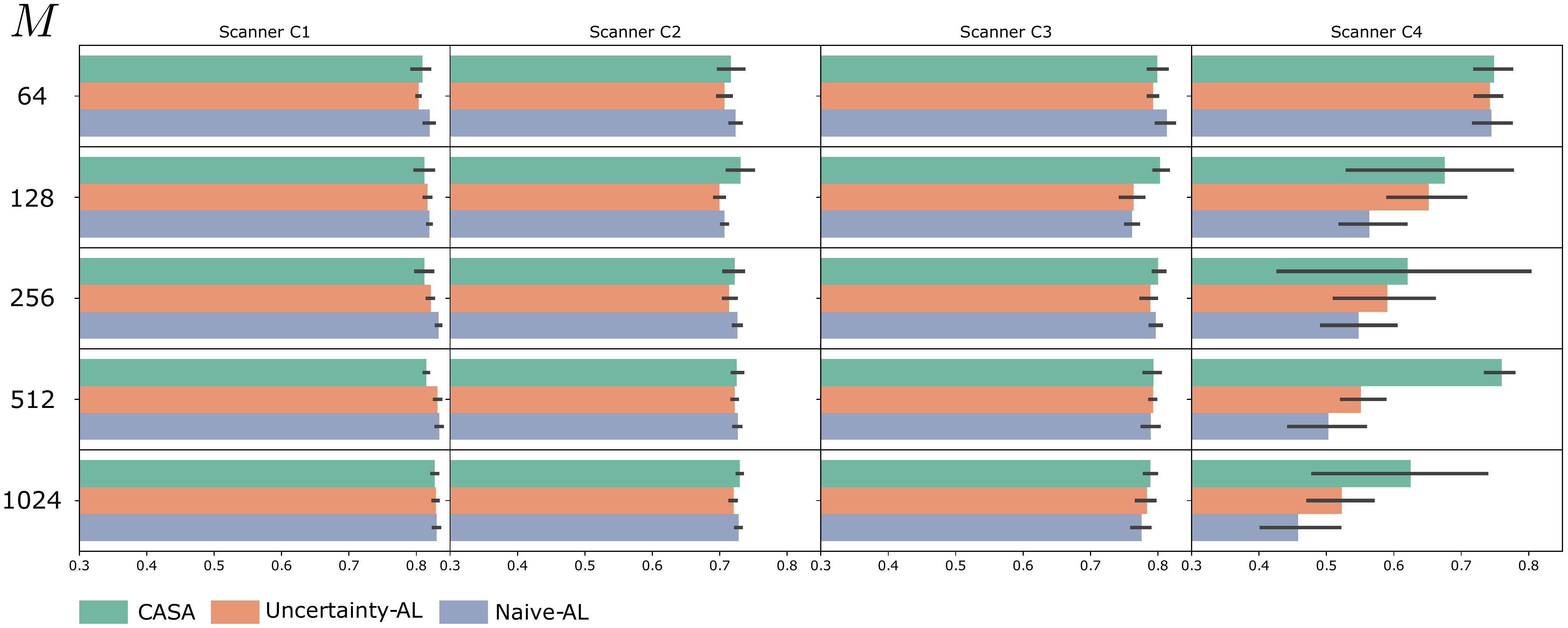}
    \caption{Influence of memory size $M$ for cardiac segmentation with labelling budget $\beta=\frac{1}{10}$ comparing CASA, uncertainty AL and naive AL. Performance was measured as mean DSC.}
    \label{fig:mem_size}
\end{figure}
The memory size $M$ influences the the adaption to new domains the the continuous data stream. In Figure \ref{fig:mem_size}, CASA, UAL and NAL are compared for $M=\langle 64, 128, 256, 512, 1024 \rangle$. For the first scanners C1 and C2 in the stream, the performance was similar across AL methods and settings of $M$. For $M=64$, CASA could not gain any improvement in comparison to UAL and NAL, meaning that the detection of pseudo-domains and balancing based on them is more useful for reasonable large memory sizes. All methods performed best for $M=64$ however, the memory on the end of the stream was primarily filled with C3 and C4 data, which would lead to forgetting effects if training continues. \revision{In addition, a performance drop for $M>=128$ compared to $M=64$ for C4 data can be observed. This is a sign that the higher the memory size, the longer it takes for all methods to adapt to new domains. At the end of the data stream training for C4 data has not saturated for a rehearsal memory of size 128 and larger. The large variation in performance of CASA on C4 data across different sizes of $M$ might be due to the early consumption of the whole labelling budget. For each independent test run, the ordering of the continuous data stream is randomly changed, some of those orderings led to CASA running out of budget before the stream ended, thus the adaptation to C4 data was not completed.}

\subsection{Evaluation of the Memory and Pseudo-Domains}
\label{sec:memory_analysis}
\begin{figure}[h]
    \centering
    \includegraphics[width=\textwidth]{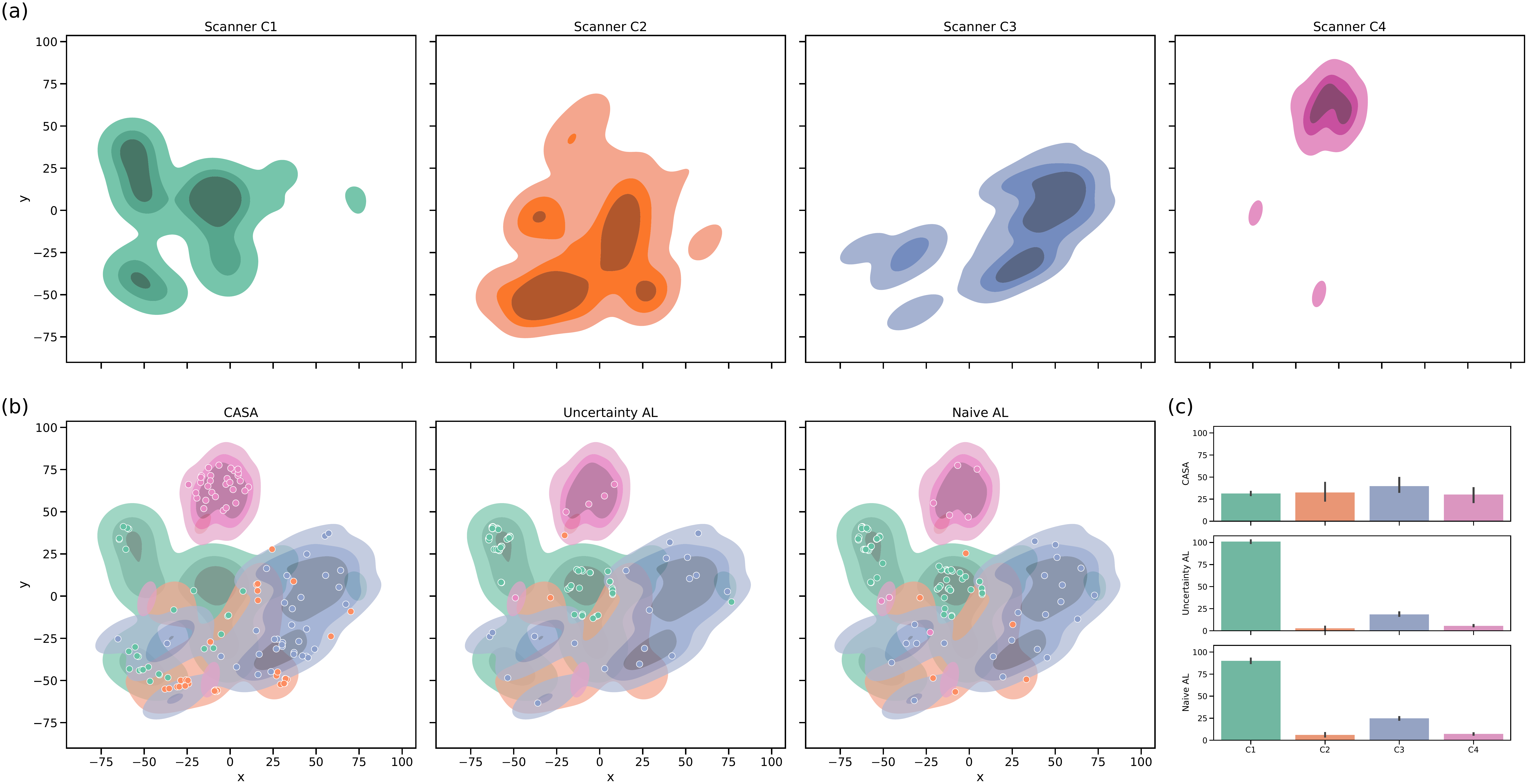}
    \caption{t-SNE visualization of style embeddings for cardiac segmentation. (a) shows the distribution of the different domains C1-C4 in the embedding space. (b) For CASA, UAL and NAL the memory elements at the end of continual training are marked in the embedding space, showing a balanced distribution for CASA. (c) Counts of elements in the rehearsal memory at the end of training for CASA, UAL and NAL.}
    \label{fig:mem_embedding}
\end{figure}
We analyzed the balancing of our memory at the end of training (with $M=128, \beta=\frac{1}{10}$) and the detection of different pseudo-domains by extracting the \textit{style embedding} for all samples in the training set (combined base and continual set). Those embeddings were mapped to two dimensions using t-SNE \citep{maaten2008visualizing} for plotting. In Figure \ref{fig:mem_embedding} (a), it is observable that different scanners are located in different areas of the embedding space. Especially scanner C4 forms a compact cluster separated from the other scanners. Furthermore, a comparison of the distribution of the samples in the rehearsal memory at the end of training (Figure \ref{fig:mem_embedding}(b)) shows that for CASA, the samples distributed over the whole embedding including scanner C4. For UAL and NAL, most samples focused on scanner C1 samples (base training scanner), and a lower number of images of later scanners were kept in memory. Note, that this does not mean that UAL and NAL labelled primarily scanner C1 samples but that those methods did not balance the rehearsal memory. So, labelled images from C2-C4 might be lost in the process of choosing what samples to keep. As shown in Appendix Figure \ref{fig:appendix_ablation_mem_methods}, those observations were stable over differnt test runs. Figure \ref{fig:mem_embedding} (c) confirms the finding. Here, the memory distribution for the compared methods over five independent runs (with different random seeds) demonstrated the capability of CASA to balance the memory across scanner, although the real domains are not known during training.

\begin{figure}[h]
    \centering
    \includegraphics[width=\textwidth]{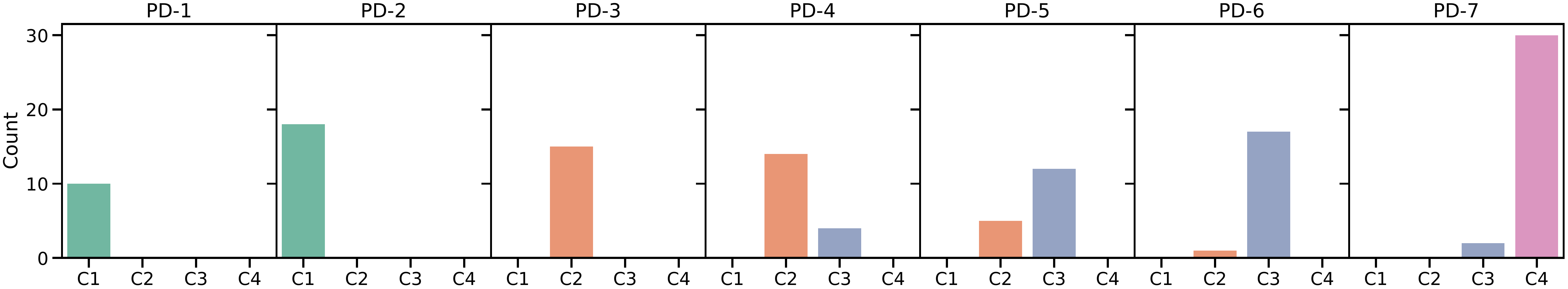}
    \caption{Distribution of images of specific scanners (C1-C4) to the discovered pseudo-domains for one run of CASA training with $M=128$, $\beta=\frac{1}{10}$. }
    \label{fig:mem_pseudodomains}
\end{figure}
Pseudo-domain discovery in CASA (with $M=128$, $\beta=\frac{1}{10}$) resulted in 6-8 pseudo-domains, showing that the definition of pseudo-domains does not exactly capture the definitions of domains in terms of scanners. In addition, the detection was influenced by the order of the continuous data stream. Figure \ref{fig:mem_pseudodomains} shows the distribution of images from certain scanners in the pseudo-domains for one training run of CASA (results for all $n=5$ independent runs are shown in Appendix Figure \ref{fig:mem_all_pd}). The first two pseudo-domains were dominated by samples from scanner C1, while the last pseudo-domain consisted mainly of scanner C4. The pseudo-domains 3-6 represented a mix of C2 and C3 data. This is consistent with Figure \ref{fig:mem_embedding} where we see that the distributions of C2 and C3 overlap while C1 and especially C4 data is more separated.

\subsection{Learning on a random stream of data}
\label{sec:random_stream}
To analyze the influence of the sequential nature of the stream, we show how CASA performs on a random stream of data, with no sequential order of the scanners used. Results are given in Table \ref{tbl:res_cardiac_random}. \revision{Note, that in this comparison adding randomness to the stream results in eliminating the domain shifts between scanners and making the samples within the stream approximately independent and identically distributed (i.i.d.).} CASA performed well on a random stream however, for scanners with few samples in the training set (Scanner C2, C4), a drop in performance was observed. This is due to the fact that the pseudo-domain detection based on the outlier memory was not as effective as on a continuous stream, where we expect a accumulation of outliers as new domains start to occur in the stream.

\begin{table}[h]
\centering
\small
\begin{tabular}{|l|l|l|l|l|}
\hline
Meth.   & C1 & C2 & C3 & C4\\
\hline
CASA - Random & $0.816\pm0.009$ & $0.719\pm0.011$ & $0.778\pm0.013$ & $0.652\pm0.078$\\
\revision{UAL - Random} & $0.820\pm0.006$ & $0.717\pm0.010$ & $0.791\pm0.006$ & $0.740\pm0.024$\\
NAL - Random &  $0.826\pm0.008$ & $0.728\pm0.007$ & $0.802\pm0.006$ & $0.745\pm0.021$\\ \hline
CASA  & $0.812\pm0.017$ & $0.731\pm0.025$ & $0.803\pm0.015$ & $0.676\pm0.158$\\ 
\revision{UAL} & $0.816\pm0.006$ & $0.700\pm0.009$ & $0.764\pm0.023$ & $0.652\pm0.078$ \\
NAL &  $0.819\pm0.003$ & $0.707\pm0.005$ & $0.761\pm0.013$ & $0.564\pm0.064$\\
\hline
\end{tabular}
\caption{Dice scores for cardiac segmentation on a random stream. CASA with $M=128$, $\beta=\frac{1}{10}$ is compared to naive active learning. $\pm$ marks the standard deviations over $n=5$ independent training runs.}
\label{tbl:res_cardiac_random}
\end{table}

NAL performed well on a random stream, outperforming NAL on a ordered stream, as well as CASA. Due to the randomness in the stream and the sampling strategy of NAL (taking every $n$-th sample), NAL learned on a diverse set of samples and managed to reach a more balanced training. \revision{Similar observations hold for uncertainty based active learning. Mixing up the order of samples in the stream leads to an earlier occurrence of data from C3 and C4, thus a better rehearsal set can be constructed with UAL.
This highlights the ability of CASA to learn under the influence of domain shifts. If no domain shifts occur in the data the specific design of the approach does not provide a benefit.}

\section{Discussion and Conclusion}

We propose a continual active learning method to adapt deep learning models to changes of medical imaging acquisition settings. By detecting novel pseudo-domains occurring in the data stream, our method is able to keep the number of required annotations low, while improving the diversity of the training set. Pseudo-domains represent groups of images with similar but new imaging characteristics. Balancing the rehearsal memory based on pseudo-domains ensures a diverse set of samples is kept for retraining on both new and preceding domains. 

Experiments showed that the proposed approach improves model accuracy in a range of different medical imaging tasks ranging from segmentation, to detection and regression. Performance of the models is improved across all domains for each task, while effectively counteracting catastrophic forgetting. Extensive experiments to gain insights on the effect of the composition of the rehearsal memory showed that CASA successfully balances training data between the real, but unknown domains.

A question that could be addressed by future research is that CASA needs to store samples that are part of the memory from preceding domains until the end of training, which could lead to data privacy concerns. A possible direction of further research is how to combine the concepts presented in this work with pseudo-rehearsal methods that do not store samples directly but rather a privacy conserving representation of the previously seen samples. \revision{In our experiments, the memory size and the labelling budget $\beta$ was fixed before training while in real world applications running for possibly unlimited time, strategies how to expand the memory to include a sufficient amount of samples to cover the whole data distribution are needed. The labelling budget $\beta$ might be increased based on the samples already observed, for example a simple strategy would be to add budget for every $n-th$ sample observed.}
\revision{Another aspect relevant for the practical implementation is that when deploying active learning approaches to real world applications, there is the need for interaction with a human annotator during the annotation. While the trained model can be used for clinical application if accuracy requirements are met, in practice active learning would indicate the necessity of annotating further cases to update the model. CASA is limiting this necessity of annotating and can be used to speed up the manual annotation process by suggesting a possible annotation for human annotators. For example when manual segmentation is required to derive a diagnosis, the proposed approach can be easily extended to enable the generation of a suggestion for a segmentation that the human annotator only has to correct, leading to time-savings compared to manual annotation from scratch.}

\revision{In experimental validation we assumed a data stream to which data from new scanners are added sequentially, i.e., one by one. However, in clinical practice, commonly multiple scanners are used in parallel, leading to possible simultaneous updates and newly entering scanners. Note that such a data stream with simultaneous additions would still be different from the randomly mixed stream analyzed in Section \ref{sec:random_stream} as multiple acquisition shifts would still appear at specific time points in clinical practice. 
In addition to multiple scanners used in parallel, it is desirable not only to include data from one hospital, but to include data from different sites. Future work is needed to explore such a multi-site and multi-stream setting. A possible approach might combine the presented approach with federated learning.
}



\acks{This work was partially supported by the Austrian Science Fund (FWF): P 35189, by the Vienna Science and Technology Fund (WWTF): LS20-065, and by Novartis Pharmaceuticals Corporation. Part of the computations for research were performed on GPUs donated by NVIDIA.}

%
\ethics{The work follows appropriate ethical standards in conducting research and writing the manuscript, following all applicable laws and regulations regarding treatment of animals or human subjects.}

\coi{M.P. and J.H. declare no conflicts of interests.
C.H.: Research Consultant for Siemens Healthineers and Bayer Healthcare, Stock holder at Hologic Inc.
H.P.: Speakers Honoraria for Boehringer Ingelheim and Roche. Received a research grant by Boehringer Ingelheim.
G.L.: Co-founder and stock holder at contextflow GmbH. Received research funding by Novartis Pharmaceuticals Corporation.}

\bibliography{references}

\begin{thebibliography}{39}
\providecommand{\natexlab}[1]{#1}
\providecommand{\url}[1]{\texttt{#1}}
\expandafter\ifx\csname urlstyle\endcsname\relax
  \providecommand{\doi}[1]{doi: #1}\else
  \providecommand{\doi}{doi: \begingroup \urlstyle{rm}\Url}\fi

\bibitem[Armato et~al.(2011)Armato, McLennan, Bidaut, McNitt-Gray, Meyer,
  Reeves, Zhao, Aberle, Henschke, Hoffman, Kazerooni, MacMahon, Van~Beek,
  Yankelevitz, Biancardi, Bland, Brown, Engelmann, Laderach, Max, Pais, Qing,
  Roberts, Smith, Starkey, Batra, Caligiuri, Farooqi, Gladish, Jude, Munden,
  Petkovska, Quint, Schwartz, Sundaram, Dodd, Fenimore, Gur, Petrick, Freymann,
  Kirby, Hughes, Vande~Casteele, Gupte, Sallam, Heath, Kuhn, Dharaiya, Burns,
  Fryd, Salganicoff, Anand, Shreter, Vastagh, Croft, and Clarke]{Armato2011}
Samuel~G. Armato, Geoffrey McLennan, Luc Bidaut, Michael~F. McNitt-Gray,
  Charles~R. Meyer, Anthony~P. Reeves, Binsheng Zhao, Denise~R. Aberle,
  Claudia~I. Henschke, Eric~A. Hoffman, Ella~A. Kazerooni, Heber MacMahon,
  Edwin~J.R. Van~Beek, David Yankelevitz, Alberto~M. Biancardi, Peyton~H.
  Bland, Matthew~S. Brown, Roger~M. Engelmann, Gary~E. Laderach, Daniel Max,
  Richard~C. Pais, David~P.Y. Qing, Rachael~Y. Roberts, Amanda~R. Smith, Adam
  Starkey, Poonam Batra, Philip Caligiuri, Ali Farooqi, Gregory~W. Gladish,
  C.~Matilda Jude, Reginald~F. Munden, Iva Petkovska, Leslie~E. Quint,
  Lawrence~H. Schwartz, Baskaran Sundaram, Lori~E. Dodd, Charles Fenimore,
  David Gur, Nicholas Petrick, John Freymann, Justin Kirby, Brian Hughes,
  Alessi Vande~Casteele, Sangeeta Gupte, Maha Sallam, Michael~D. Heath,
  Michael~H. Kuhn, Ekta Dharaiya, Richard Burns, David~S. Fryd, Marcos
  Salganicoff, Vikram Anand, Uri Shreter, Stephen Vastagh, Barbara~Y. Croft,
  and Laurence~P. Clarke.
\newblock {The Lung Image Database Consortium (LIDC) and Image Database
  Resource Initiative (IDRI): A completed reference database of lung nodules on
  CT scans}.
\newblock \emph{Medical Physics}, 38\penalty0 (2):\penalty0 915--931, 2011.
\newblock ISSN 00942405.
\newblock \doi{10.1118/1.3528204}.

\bibitem[Beer et~al.(2020)Beer, Tustison, Cook, Davatzikos, Sheline, Shinohara,
  and Linn]{Beer2020LongitudinalData}
Joanne~C. Beer, Nicholas~J. Tustison, Philip~A. Cook, Christos Davatzikos,
  Yvette~I. Sheline, Russell~T. Shinohara, and Kristin~A. Linn.
\newblock {Longitudinal ComBat: A method for harmonizing longitudinal
  multi-scanner imaging data}.
\newblock \emph{NeuroImage}, 220, 10 2020.
\newblock ISSN 10959572.
\newblock \doi{10.1016/j.neuroimage.2020.117129}.

\bibitem[Bobu et~al.(2018)Bobu, Tzeng, Hoffman, and
  Darrell]{Bobu2018AdaptingDomains}
Andreea Bobu, Eric Tzeng, Judy Hoffman, and Trevor Darrell.
\newblock {Adapting to continously shifting domains}.
\newblock In \emph{ICLR Workshop}, 2018.

\bibitem[Budd et~al.(2019)Budd, Robinson, and Kainz]{Budd2019}
Samuel Budd, Emma~C Robinson, and Bernhard Kainz.
\newblock {A Survey on Active Learning and Human-in-the-Loop Deep Learning for
  Medical Image Analysis}.
\newblock 2019.
\newblock URL \url{http://arxiv.org/abs/1910.02923}.

\bibitem[Campello et~al.(2021)Campello, Gkontra, Izquierdo, Martin-Isla,
  Sojoudi, Full, Maier-Hein, Zhang, He, Ma, Parreno, Albiol, Kong, Shadden,
  Acero, Sundaresan, Saber, Elattar, Li, Menze, Khader, Haarburger, Scannell,
  Veta, Carscadden, Punithakumar, Liu, Tsaftaris, Huang, Yang, Li, Zhuang,
  Vilades, Descalzo, Guala, La~Mura, Friedrich, Garg, Lebel, Henriques,
  Karakas, Cavus, Petersen, Escalera, Segui, Rodriguez-Palomares, and
  Lekadir]{Campello2021Multi-CentreChallenge}
Victor~M. Campello, Polyxeni Gkontra, Cristian Izquierdo, Carlos Martin-Isla,
  Alireza Sojoudi, Peter~M. Full, Klaus Maier-Hein, Yao Zhang, Zhiqiang He, Jun
  Ma, Mario Parreno, Alberto Albiol, Fanwei Kong, Shawn~C. Shadden,
  Jorge~Corral Acero, Vaanathi Sundaresan, Mina Saber, Mustafa Elattar, Hongwei
  Li, Bjoern Menze, Firas Khader, Christoph Haarburger, Cian~M. Scannell, Mitko
  Veta, Adam Carscadden, Kumaradevan Punithakumar, Xiao Liu, Sotirios~A.
  Tsaftaris, Xiaoqiong Huang, Xin Yang, Lei Li, Xiahai Zhuang, David Vilades,
  Martin~L. Descalzo, Andrea Guala, Lucia La~Mura, Matthias~G. Friedrich, Ria
  Garg, Julie Lebel, Filipe Henriques, Mahir Karakas, Ersin Cavus, Steffen~E.
  Petersen, Sergio Escalera, Santi Segui, Jose~F. Rodriguez-Palomares, and
  Karim Lekadir.
\newblock {Multi-Centre, Multi-Vendor and Multi-Disease Cardiac Segmentation:
  The M{\&}amp;Ms Challenge}.
\newblock \emph{IEEE Transactions on Medical Imaging}, pages 1--1, 6 2021.
\newblock \doi{10.1109/TMI.2021.3090082}.

\bibitem[Castro et~al.(2020)Castro, Walker, and Glocker]{Castro2020}
Daniel~C. Castro, Ian Walker, and Ben Glocker.
\newblock {Causality matters in medical imaging}.
\newblock \emph{Nature Communications}, 11\penalty0 (1):\penalty0 1--10, 2020.
\newblock ISSN 20411723.
\newblock \doi{10.1038/s41467-020-17478-w}.
\newblock URL \url{http://dx.doi.org/10.1038/s41467-020-17478-w}.

\bibitem[Delange et~al.(2021)Delange, Aljundi, Masana, Parisot, Jia, Leonardis,
  Slabaugh, and Tuytelaars]{Delange2021ATasks}
Matthias Delange, Rahaf Aljundi, Marc Masana, Sarah Parisot, Xu~Jia, Ales
  Leonardis, Greg Slabaugh, and Tinne Tuytelaars.
\newblock {A continual learning survey: Defying forgetting in classification
  tasks}.
\newblock \emph{IEEE Transactions on Pattern Analysis and Machine
  Intelligence}, pages 1--1, 2021.
\newblock ISSN 0162-8828.
\newblock \doi{10.1109/TPAMI.2021.3057446}.
\newblock URL \url{https://ieeexplore.ieee.org/document/9349197/}.

\bibitem[Dinsdale et~al.(2020)Dinsdale, Jenkinson, and Namburete]{Dinsdale2020}
Nicola~K. Dinsdale, Mark Jenkinson, and Ana~I.L. Namburete.
\newblock {Unlearning Scanner Bias for MRI Harmonisation in Medical Image
  Segmentation}.
\newblock \emph{Communications in Computer and Information Science}, 1248
  CCIS:\penalty0 15--25, 2020.
\newblock ISSN 18650937.
\newblock \doi{10.1007/978-3-030-52791-4{\_}2}.

\bibitem[Everingham et~al.(2010)Everingham, Van~Gool, Williams, Winn, and
  Zisserman]{Everingham2010}
Mark Everingham, Luc Van~Gool, Christopher K.~I. Williams, John Winn, and
  Andrew Zisserman.
\newblock {The Pascal Visual Object Classes (VOC) Challenge}.
\newblock \emph{International Journal of Computer Vision}, 88\penalty0
  (2):\penalty0 303--338, 6 2010.
\newblock ISSN 0920-5691.
\newblock \doi{10.1007/s11263-009-0275-4}.
\newblock URL \url{http://link.springer.com/10.1007/s11263-009-0275-4}.

\bibitem[Fortin et~al.(2018)Fortin, Cullen, Sheline, Taylor, Aselcioglu, Cook,
  Adams, Cooper, Fava, McGrath, McInnis, Phillips, Trivedi, Weissman, and
  Shinohara]{Fortin2018HarmonizationSites}
Jean~Philippe Fortin, Nicholas Cullen, Yvette~I. Sheline, Warren~D. Taylor,
  Irem Aselcioglu, Philip~A. Cook, Phil Adams, Crystal Cooper, Maurizio Fava,
  Patrick~J. McGrath, Melvin McInnis, Mary~L. Phillips, Madhukar~H. Trivedi,
  Myrna~M. Weissman, and Russell~T. Shinohara.
\newblock {Harmonization of cortical thickness measurements across scanners and
  sites}.
\newblock \emph{NeuroImage}, 167:\penalty0 104--120, 2 2018.
\newblock ISSN 10959572.
\newblock \doi{10.1016/j.neuroimage.2017.11.024}.

\bibitem[Gal and Ghahramani(2016)]{Gal2016}
Yarin Gal and Zoubin Ghahramani.
\newblock {Dropout as a Bayesian Approximation: Representing Model Uncertainty
  in Deep Learning Zoubin Ghahramani}.
\newblock In \emph{Proceedings of The 33rd International Conference on Machine
  Learning}, pages 1050--1059, 2016.

\bibitem[Gatys et~al.(2016)Gatys, Ecker, and Bethge]{Gatys2016}
Leon Gatys, Alexander Ecker, and Matthias Bethge.
\newblock {A Neural Algorithm of Artistic Style}.
\newblock \emph{Journal of Vision}, 16\penalty0 (12):\penalty0 326, 2016.
\newblock ISSN 1534-7362.
\newblock \doi{10.1167/16.12.326}.

\bibitem[Glocker et~al.(2019)Glocker, Robinson, Castro, Dou, and
  Konukoglu]{Glocker2019MachineEffects}
Ben Glocker, Robert Robinson, Daniel~C. Castro, Qi~Dou, and Ender Konukoglu.
\newblock {Machine Learning with Multi-Site Imaging Data: An Empirical Study on
  the Impact of Scanner Effects}.
\newblock \emph{arXiv Preprint}, 2019.
\newblock URL \url{http://arxiv.org/abs/1910.04597}.

\bibitem[Gonzalez et~al.(2020)Gonzalez, Sakas, and Mukhopadhyay]{Gonzalez2020}
Camila Gonzalez, Georgios Sakas, and Anirban Mukhopadhyay.
\newblock {What is Wrong with Continual Learning in Medical Image
  Segmentation?}
\newblock \emph{arXiv Preprint}, 10 2020.
\newblock URL \url{http://arxiv.org/abs/2010.11008}.

\bibitem[Guan and Liu(2021)]{Guan2021DomainSurvey}
Hao Guan and Mingxia Liu.
\newblock {Domain Adaptation for Medical Image Analysis: A Survey}.
\newblock 2 2021.
\newblock URL \url{http://arxiv.org/abs/2102.09508}.

\bibitem[Hofmanninger et~al.(2020)Hofmanninger, Perkonigg, Brink, Pianykh,
  Herold, and Langs]{Hofmanninger2020a}
Johannes Hofmanninger, Matthias Perkonigg, James~A. Brink, Oleg Pianykh,
  Christian Herold, and Georg Langs.
\newblock {Dynamic Memory to Alleviate Catastrophic Forgetting in Continuous
  Learning Settings}.
\newblock \emph{Lecture Notes in Computer Science (including subseries Lecture
  Notes in Artificial Intelligence and Lecture Notes in Bioinformatics)}, 12262
  LNCS:\penalty0 359--368, 2020.
\newblock ISSN 16113349.
\newblock \doi{10.1007/978-3-030-59713-9{\_}35}.

\bibitem[Karani et~al.(2018)Karani, Chaitanya, Baumgartner, and
  Konukoglu]{Karani2018}
Neerav Karani, Krishna Chaitanya, Christian Baumgartner, and Ender Konukoglu.
\newblock {A lifelong learning approach to brain MR segmentation across
  scanners and protocols}.
\newblock In \emph{Medical Image Computing and Computer-Assisted Intervention
  (MICCAI)}, volume 11070 LNCS, pages 476--484. Springer, Cham, 2018.
\newblock ISBN 9783030009274.
\newblock \doi{10.1007/978-3-030-00928-1{\_}54}.

\bibitem[LaMontagne et~al.(2019)LaMontagne, Benzinger, Morris, Keefe, Hornbeck,
  Xiong, Grant, Hassenstab, Moulder, Vlassenko, Raichle, Cruchaga, and
  Marcus]{LaMontagne2019}
Pamela~J LaMontagne, Tammie L~S Benzinger, John~C Morris, Sarah Keefe, Russ
  Hornbeck, Chengjie Xiong, Elizabeth Grant, Jason Hassenstab, Krista Moulder,
  Andrei~G Vlassenko, Marcus~E Raichle, Carlos Cruchaga, and Daniel Marcus.
\newblock {OASIS-3: Longitudinal Neuroimaging, Clinical, and Cognitive Dataset
  for Normal Aging and Alzheimer Disease}.
\newblock \emph{medRxiv}, page 2019.12.13.19014902, 2019.
\newblock \doi{10.1101/2019.12.13.19014902}.
\newblock URL
  \url{http://medrxiv.org/content/early/2019/12/15/2019.12.13.19014902.abstract}.

\bibitem[Lao et~al.(2020)Lao, Jiang, Havaei, and Bengio]{Lao2020}
Qicheng Lao, Xiang Jiang, Mohammad Havaei, and Yoshua Bengio.
\newblock {Continuous Domain Adaptation with Variational Domain-Agnostic
  Feature Replay}.
\newblock 2020.
\newblock URL \url{http://arxiv.org/abs/2003.04382}.

\bibitem[Lenga et~al.(2020)Lenga, Schulz, and Saalbach]{Lenga2020}
Matthias Lenga, Heinrich Schulz, and Axel Saalbach.
\newblock {Continual Learning for Domain Adaptation in Chest X-ray
  Classification}.
\newblock In \emph{Conference on Medical Imaging with Deep Learning (MIDL)},
  2020.
\newblock URL \url{http://arxiv.org/abs/2001.05922}.

\bibitem[Liu et~al.(2008)Liu, Ting, and Zhou]{Liu2008}
Tony~Fei Liu, Ming~Kai Ting, and Zhi-Hua Zhou.
\newblock {Isolation Forest}.
\newblock In \emph{International Conference on Data Mining}, 2008.

\bibitem[Lopez-Paz and Ranzato(2017)]{Lopez-Paz2017}
David Lopez-Paz and Marc'Aurelio Ranzato.
\newblock {Gradient episodic memory for continual learning}.
\newblock \emph{Advances in Neural Information Processing Systems}, pages
  6468--6477, 2017.
\newblock ISSN 10495258.

\bibitem[Maaten and Hinton(2008)]{maaten2008visualizing}
Laurens van~der Maaten and Geoffrey Hinton.
\newblock {Visualizing data using t-SNE}.
\newblock \emph{Journal of machine learning research}, 9\penalty0
  (Nov):\penalty0 2579--2605, 2008.

\bibitem[McCloskey and Cohen(1989)]{McCloskey1989}
Michael McCloskey and Neal~J. Cohen.
\newblock {Catastrophic Interference in Connectionist Networks: The Sequential
  Learning Problem}.
\newblock \emph{Psychology of Learning and Motivation - Advances in Research
  and Theory}, 24\penalty0 (C):\penalty0 109--165, 1989.
\newblock ISSN 00797421.
\newblock \doi{10.1016/S0079-7421(08)60536-8}.

\bibitem[Ozdemir et~al.(2018)Ozdemir, Fuernstahl, and Goksel]{Ozdemir2018}
Firat Ozdemir, Philipp Fuernstahl, and Orcun Goksel.
\newblock {Learn the New, Keep the Old: Extending Pretrained Models with New
  Anatomy and Images}.
\newblock \emph{Lecture Notes in Computer Science (including subseries Lecture
  Notes in Artificial Intelligence and Lecture Notes in Bioinformatics)}, 11073
  LNCS:\penalty0 361--369, 2018.
\newblock ISSN 16113349.
\newblock \doi{10.1007/978-3-030-00937-3{\_}42}.

\bibitem[{\"{O}}zg{\"{u}}n et~al.(2020){\"{O}}zg{\"{u}}n, Rickmann, Roy, and
  Wachinger]{Ozgun2020}
Sinan {\"{O}}zg{\"{u}}n, Anne-Marie Rickmann, Abhijit~Guha Roy, and Christian
  Wachinger.
\newblock {Importance Driven Continual Learning for Segmentation Across
  Domains}.
\newblock Number~Cl, pages 423--433. 2020.
\newblock \doi{10.1007/978-3-030-59861-7{\_}43}.
\newblock URL \url{https://link.springer.com/10.1007/978-3-030-59861-7_43}.

\bibitem[Parisi et~al.(2019)Parisi, Kemker, Part, Kanan, and
  Wermter]{Parisi2019ContinualReview}
German~I. Parisi, Ronald Kemker, Jose~L. Part, Christopher Kanan, and Stefan
  Wermter.
\newblock {Continual lifelong learning with neural networks: A review}.
\newblock \emph{Neural Networks}, 113:\penalty0 54--71, 5 2019.
\newblock ISSN 08936080.
\newblock \doi{10.1016/j.neunet.2019.01.012}.
\newblock URL
  \url{https://linkinghub.elsevier.com/retrieve/pii/S0893608019300231}.

\bibitem[Pedrosa et~al.(2019)Pedrosa, Aresta, Ferreira, Rodrigues,
  Leit{\~{a}}o, Carvalho, Rebelo, Negr{\~{a}}o, Ramos, Cunha, and
  Campilho]{Pedrosa2019}
João Pedrosa, Guilherme Aresta, Carlos Ferreira, Márcio Rodrigues, Patrícia
  Leit{\~{a}}o, André~Silva Carvalho, João Rebelo, Eduardo Negr{\~{a}}o,
  Isabel Ramos, António Cunha, and Aurélio Campilho.
\newblock {LNDb: A lung nodule database on computed tomography}.
\newblock \emph{arXiv}, pages 1--12, 2019.
\newblock ISSN 23318422.

\bibitem[Perkonigg et~al.(2021{\natexlab{a}})Perkonigg, Hofmanninger, Herold,
  Brink, Pianykh, Prosch, and Langs]{Perkonigg2021DynamicImaging}
Matthias Perkonigg, Johannes Hofmanninger, Christian~J. Herold, James~A. Brink,
  Oleg Pianykh, Helmut Prosch, and Georg Langs.
\newblock {Dynamic memory to alleviate catastrophic forgetting in continual
  learning with medical imaging}.
\newblock \emph{Nature Communications}, 12\penalty0 (1):\penalty0 5678, 12
  2021{\natexlab{a}}.
\newblock ISSN 2041-1723.
\newblock \doi{10.1038/s41467-021-25858-z}.
\newblock URL \url{https://www.nature.com/articles/s41467-021-25858-z}.

\bibitem[Perkonigg et~al.(2021{\natexlab{b}})Perkonigg, Hofmanninger, and
  Langs]{Perkonigg2021ContinualAcquisition}
Matthias Perkonigg, Johannes Hofmanninger, and Georg Langs.
\newblock {Continual Active Learning for Efficient Adaptation of Machine
  Learning Models to Changing Image Acquisition}.
\newblock In \emph{Advances in Information Processing in Medical Imaging,
  IPMI}, 2021{\natexlab{b}}.

\bibitem[Pianykh et~al.(2020)Pianykh, Langs, Dewey, Enzmann, Herold,
  Schoenberg, and Brink]{Pianykh2020}
Oleg~S. Pianykh, Georg Langs, Marc Dewey, Dieter~R. Enzmann, Christian~J.
  Herold, Stefan~O. Schoenberg, and James~A. Brink.
\newblock {Continuous learning AI in radiology: Implementation principles and
  early applications}.
\newblock \emph{Radiology}, 297\penalty0 (1):\penalty0 6--14, 2020.
\newblock ISSN 15271315.
\newblock \doi{10.1148/radiol.2020200038}.

\bibitem[Prayer et~al.(2021)Prayer, Hofmanninger, Weber, Kifjak, Willenpart,
  Pan, R{\"{o}}hrich, Langs, and Prosch]{Prayer2021VariabilityStudy}
Florian Prayer, Johannes Hofmanninger, Michael Weber, Daria Kifjak, Alexander
  Willenpart, Jeanny Pan, Sebastian R{\"{o}}hrich, Georg Langs, and Helmut
  Prosch.
\newblock {Variability of computed tomography radiomics features of fibrosing
  interstitial lung disease: A test-retest study}.
\newblock \emph{Methods}, 188:\penalty0 98--104, 2021.

\bibitem[Ren et~al.(2017)Ren, He, Girshick, and Sun]{Ren2017}
Shaoqing Ren, Kaiming He, Ross Girshick, and Jian Sun.
\newblock {Faster R-CNN: Towards Real-Time Object Detection with Region
  Proposal Networks}.
\newblock \emph{IEEE Transactions on Pattern Analysis and Machine
  Intelligence}, 39\penalty0 (6):\penalty0 1137--1149, 2017.
\newblock ISSN 01628828.
\newblock \doi{10.1109/TPAMI.2016.2577031}.

\bibitem[Ronneberger et~al.(2015)Ronneberger, Fischer, and
  Brox]{Ronneberger2015}
Olaf Ronneberger, Philipp Fischer, and Thomas Brox.
\newblock {U-Net: Convolutional Networks for Biomedical Image Segmentation}.
\newblock pages 1--8, 2015.
\newblock ISSN 16113349.
\newblock \doi{10.1007/978-3-319-24574-4{\_}28}.
\newblock URL \url{http://arxiv.org/abs/1505.04597}.

\bibitem[Setio et~al.(2017)Setio, Traverso, de~Bel, Berens, Bogaard, Cerello,
  Chen, Dou, Fantacci, Geurts, Gugten, Heng, Jansen, de~Kaste, Kotov, Lin,
  Manders, S{\'{o}}{\~{n}}ora-Mengana, Garc{\'{i}}a-Naranjo, Papavasileiou,
  Prokop, Saletta, Schaefer-Prokop, Scholten, Scholten, Snoeren, Torres,
  Vandemeulebroucke, Walasek, Zuidhof, Ginneken, and Jacobs]{Setio2017}
Arnaud Arindra~Adiyoso Setio, Alberto Traverso, Thomas de~Bel, Moira~S.N.
  Berens, Cas van~den Bogaard, Piergiorgio Cerello, Hao Chen, Qi~Dou,
  Maria~Evelina Fantacci, Bram Geurts, Robbert van~der Gugten, Pheng~Ann Heng,
  Bart Jansen, Michael~M.J. de~Kaste, Valentin Kotov, Jack Yu-Hung Lin,
  Jeroen~T.M.C. Manders, Alexander S{\'{o}}{\~{n}}ora-Mengana, Juan~Carlos
  Garc{\'{i}}a-Naranjo, Evgenia Papavasileiou, Mathias Prokop, Marco Saletta,
  Cornelia~M Schaefer-Prokop, Ernst~T. Scholten, Luuk Scholten, Miranda~M.
  Snoeren, Ernesto~Lopez Torres, Jef Vandemeulebroucke, Nicole Walasek,
  Guido~C.A. Zuidhof, Bram~van Ginneken, and Colin Jacobs.
\newblock {Validation, comparison, and combination of algorithms for automatic
  detection of pulmonary nodules in computed tomography images: The LUNA16
  challenge}.
\newblock \emph{Medical Image Analysis}, 42:\penalty0 1--13, 12 2017.
\newblock ISSN 13618415.
\newblock \doi{10.1016/j.media.2017.06.015}.
\newblock URL
  \url{https://linkinghub.elsevier.com/retrieve/pii/S1361841517301020}.

\bibitem[Smailagic et~al.(2020)Smailagic, Costa, Gaudio, Khandelwal,
  Mirshekari, Fagert, Walawalkar, Xu, Galdran, Zhang, Campilho, and
  Noh]{Smailagic2020}
Asim Smailagic, Pedro Costa, Alex Gaudio, Kartik Khandelwal, Mostafa
  Mirshekari, Jonathon Fagert, Devesh Walawalkar, Susu Xu, Adrian Galdran, Pei
  Zhang, Aurélio Campilho, and Hae~Young Noh.
\newblock {O-MedAL: Online active deep learning for medical image analysis}.
\newblock \emph{Wiley Interdisciplinary Reviews: Data Mining and Knowledge
  Discovery}, 10\penalty0 (4):\penalty0 1--15, 2020.
\newblock ISSN 19424795.
\newblock \doi{10.1002/widm.1353}.

\bibitem[Wu et~al.(2019)Wu, Wang, Gonzalez, Goldstein, and Davis]{Wu2019}
Zuxuan Wu, Xin Wang, Joseph Gonzalez, Tom Goldstein, and Larry Davis.
\newblock {ACE: Adapting to changing environments for semantic segmentation}.
\newblock \emph{Proceedings of the IEEE International Conference on Computer
  Vision}, pages 2121--2130, 2019.
\newblock ISSN 15505499.
\newblock \doi{10.1109/ICCV.2019.00221}.

\bibitem[Zhou et~al.(2020)Zhou, Sodha, Pang, Gotway, and Liang]{Zhou2020}
Zongwei Zhou, Vatsal Sodha, Jiaxuan Pang, Michael~B. Gotway, and Jianming
  Liang.
\newblock {Models Genesis}.
\newblock \emph{Medical Image Analysis}, page 101840, 2020.
\newblock ISSN 13618415.
\newblock \doi{10.1016/j.media.2020.101840}.

\bibitem[Zhou et~al.(2021)Zhou, Shin, Gurudu, Gotway, and
  Liang]{Zhou2021ActiveEfforts}
Zongwei Zhou, Jae~Y. Shin, Suryakanth~R. Gurudu, Michael~B. Gotway, and
  Jianming Liang.
\newblock {Active, continual fine tuning of convolutional neural networks for
  reducing annotation efforts}.
\newblock \emph{Medical Image Analysis}, 71, 7 2021.
\newblock ISSN 13618423.
\newblock \doi{10.1016/j.media.2021.101997}.

\end{thebibliography}

\newpage
\appendix 
\section*{Appendix A.}

\begin{figure}[h]
    \centering
    \includegraphics[width=\textwidth]{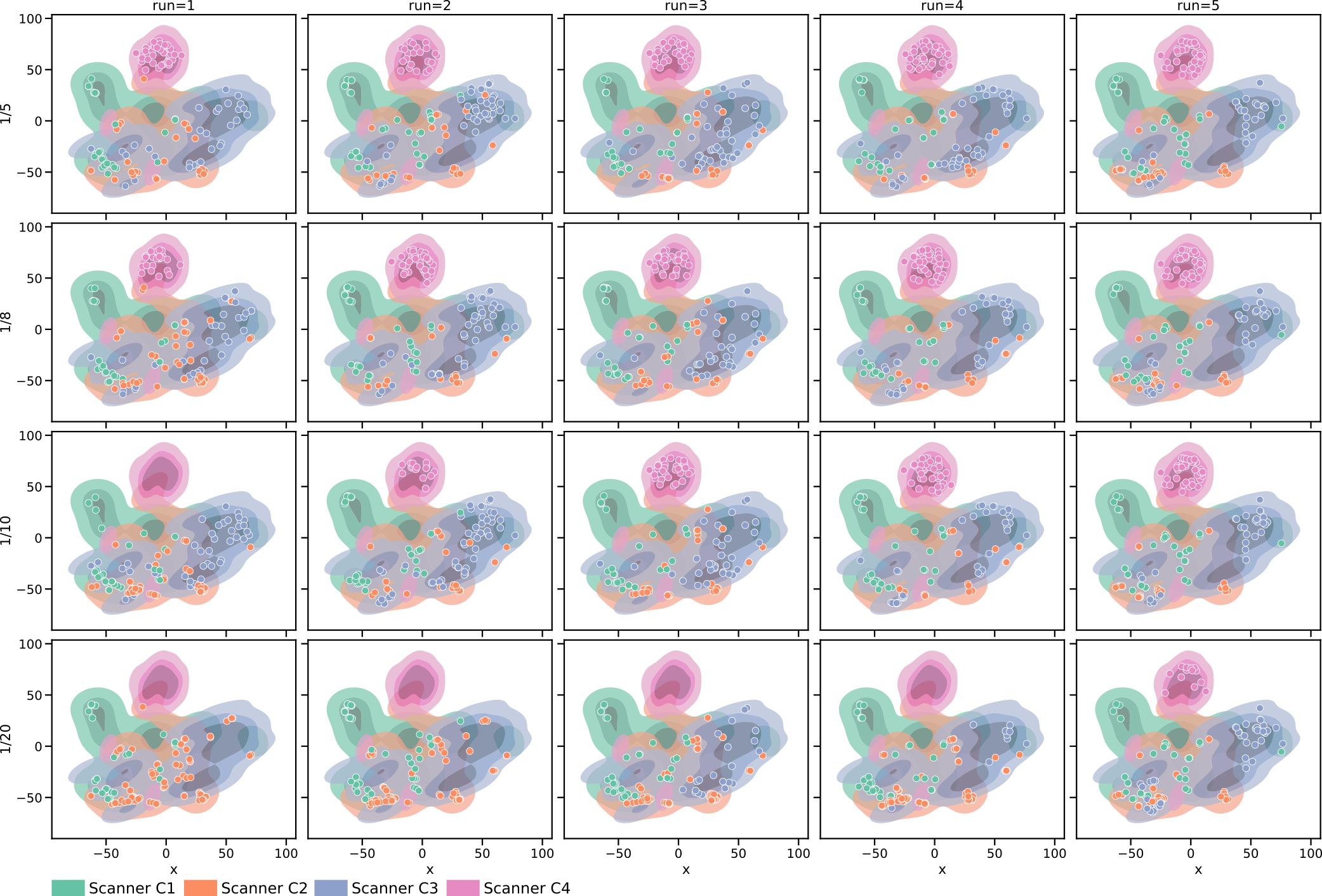}
    \caption{Style embeddings of CASA training memories for different runs with different labelling budgets $\beta$.}
    \label{fig:appendix_ablation_mem_budgets}
\end{figure}

\begin{figure}[h]
    \centering
    \includegraphics[width=\textwidth]{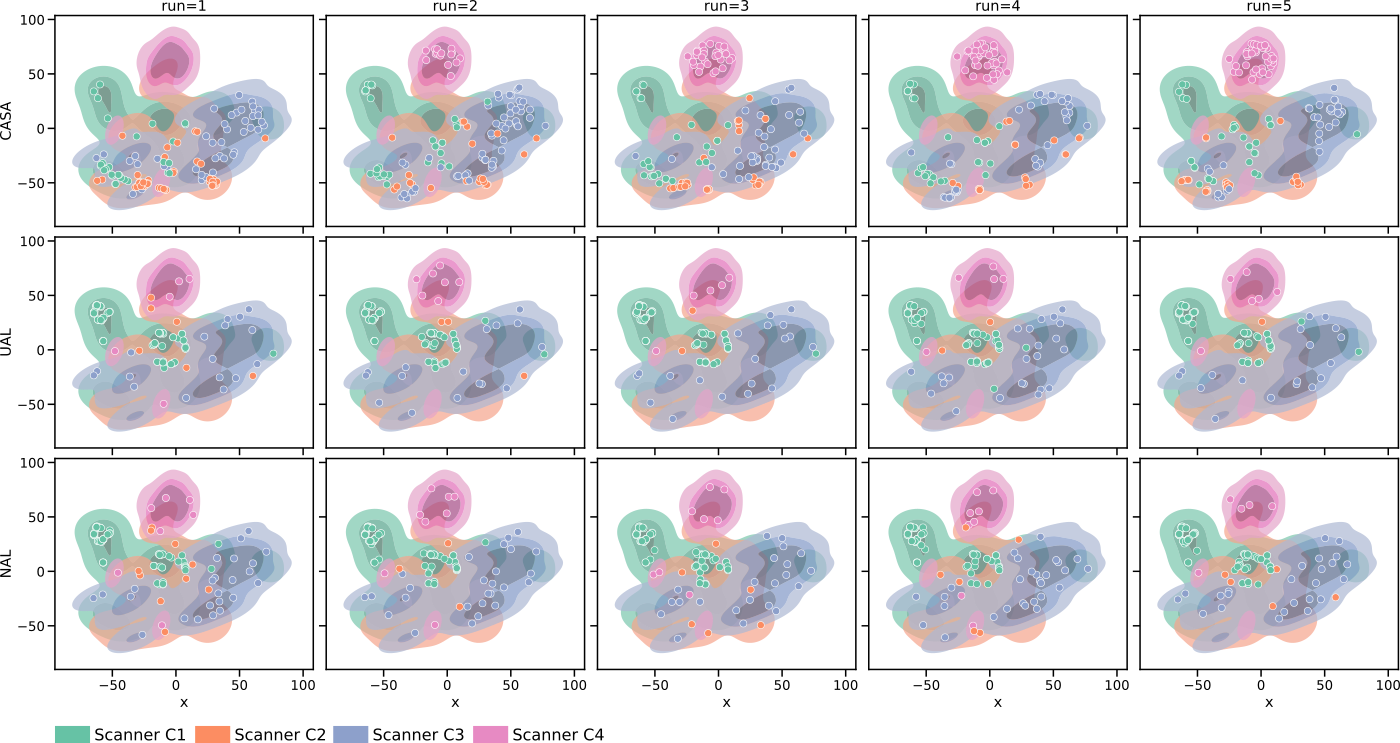}
    \caption{Style embeddings of training memories for different runs of CASA, UAL and NAL.}
    \label{fig:appendix_ablation_mem_methods}
\end{figure}

\begin{figure}[h]
    \centering
    \includegraphics[width=\textwidth]{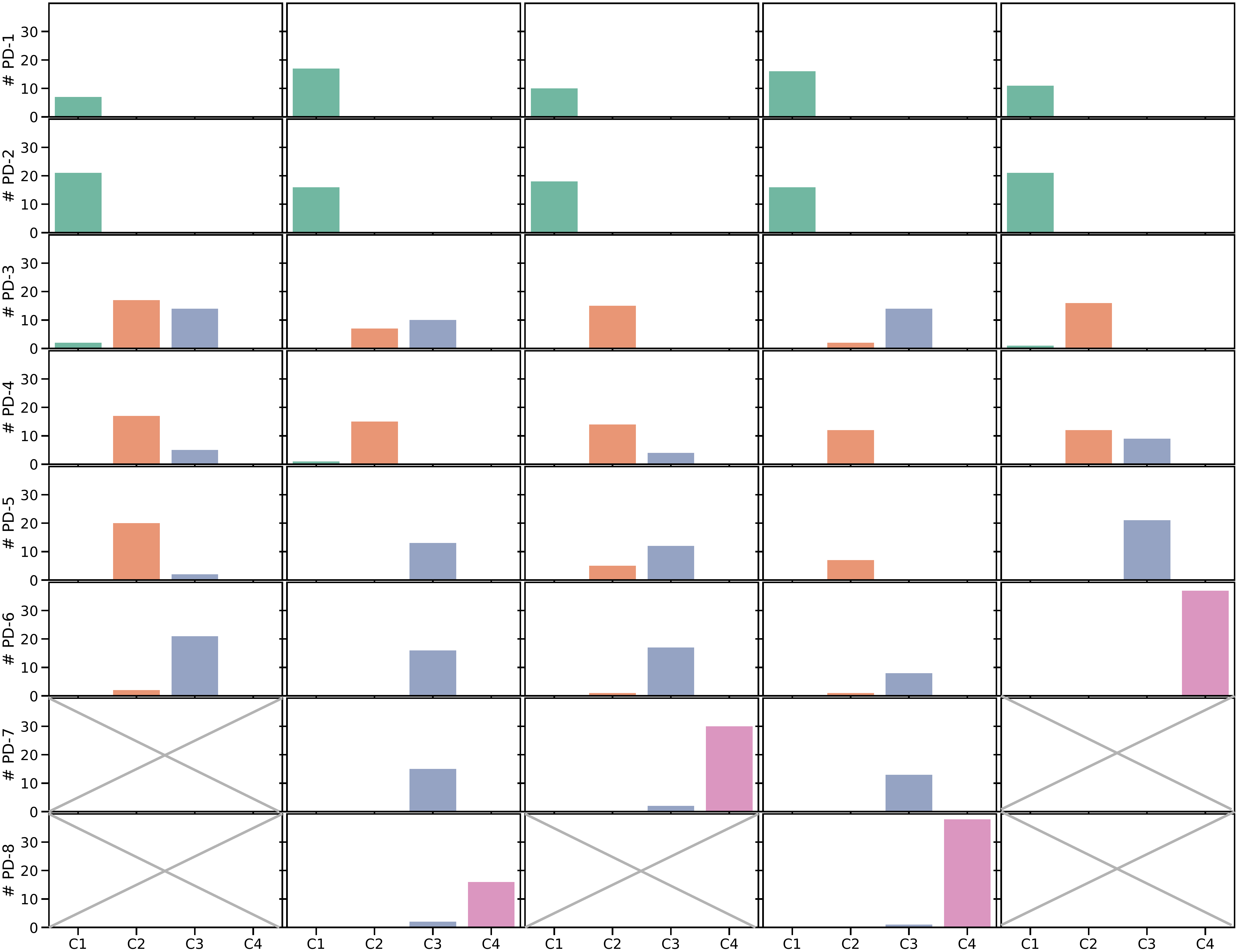}
    \caption{Distribution of images acquired with a specific scanner (C1-C4) to the discovered pseudo-domains for five runs of CASA training with $M=128$, $\beta=\frac{1}{10}$.}
    \label{fig:mem_all_pd}
\end{figure}

\noindent

\end{document}